\documentclass[11pt]{article} 
\usepackage[margin=1in]{geometry}
\usepackage{times}
\usepackage{hyperref}
\usepackage{url}

\usepackage{graphicx}
\usepackage{amsmath,amssymb}
\usepackage{booktabs}
\usepackage{multirow}
\usepackage{tabularx}
\usepackage{framed}
\usepackage{placeins}
\usepackage{tikz}
\usepackage{natbib}
\usetikzlibrary{arrows.meta,positioning}
\newcommand{\tablecaptionspace}{\vspace{\baselineskip}}

\newcounter{algline}
\newcommand{\algln}{\stepcounter{algline}\thealgline:~}
\newenvironment{algbox}[2]{%
  \setcounter{algline}{0}%
  \begin{figure}[t]
  \begin{framed}
  \small
  \textbf{Algorithm #1} #2\par\vspace{2pt}\hrule\vspace{4pt}
}{%
  \end{framed}
  \end{figure}
}

\newenvironment{algboxhere}[2]{%
  \setcounter{algline}{0}%
  \begin{framed}
  \small
  \setlength{\parskip}{1.5pt}%
  \textbf{Algorithm #1} #2\par\vspace{2pt}\hrule\vspace{4pt}
}{%
  \end{framed}
}

\title{The Arbitrary-Placement Problem in Entropy-Minimizing Selection, and a Residual-Entropy Formulation}

\author{
\begin{tabular}{cc}
\begin{minipage}{0.45\textwidth}
\centering
Alyssa H. Shin\\
\small Division of Biology and Biological Engineering\\
\small California Institute of Technology\\
\texttt{\small ahshin@caltech.edu}
\end{minipage}
&
\begin{minipage}{0.45\textwidth}
\centering
Claire H. Shin\\
\small Robert Frederick Smith School of Chemical\\
\small and Biomolecular Engineering\\
\small Cornell University\\
\texttt{\small chs237@cornell.edu}
\end{minipage}
\end{tabular}
}

\date{}

\newcommand{\R}{\mathbb{R}}

\begin{document}

\maketitle

\begin{abstract}
Entropy-based selection objectives suffer from a fundamental degeneracy: minimizing Shannon entropy $H(p_A)$ rewards confident selection regardless of whether the selected candidate is informative. We address this limitation with the residual entropy
\[
D = H(p_A) - H(p_\beta),
\]
where $p_\beta$ is induced directly by candidate trust weights before they are aggregated into scores. We prove the exact identity
\[
D = -\mathrm{KL}(p_A\Vert p_\beta) - \Delta,
\]
where $\Delta$ measures whether the score-induced distribution and the underlying trust profile favor the same candidates. The boundary cases establish basic safety: under uniform trust, $D\leq0$ automatically, so an equal-trust, non-starving state is never penalized, while at any one-hot limit, $D\to0$ regardless of the selected candidate. Our main results address the intermediate regime where selection occurs. We prove that $D\leq0$ when the ordering of candidates by trust agrees pairwise with their ordering by informativeness, giving a checkable pairwise condition for safe informative selection. We further derive a tighter certificate based on the margin of the leading candidate over its competitors, which determines when that candidate can be selected while preserving $D\leq0$. These results are independent of the candidate-scoring function and apply to both stationary and dynamically changing information. Experiments with a gradient-based mixture-of-experts router confirm that the ordering conditions can hold during real optimization and show that correct ordering improves downstream performance when candidates are non-interchangeable and selections are used directly rather than averaged. Beyond routing, margin-based reweighting matches or outperforms fixed-strength baselines in a class-imbalance task, while informative selection in a production video-prediction system reduces MSE by approximately 20\% and transfers to a related species. Residual entropy, therefore, provides a safety criterion for selection and a usable signal for deciding when that selection is informative.
\end{abstract}

\section{Introduction}

Many selection systems form distributions over candidates and use Shannon entropy minimization to encourage confident decisions. Lower entropy implies a more confident and potentially more useful choice. However, $H(p)$ is minimized by any sufficiently concentrated distribution, regardless of which candidate receives the mass. Entropy alone, therefore, cannot distinguish an informative choice from an arbitrary confident one. Mixture-of-experts (MoE) routing provides a concrete setting in which this degeneracy manifests: without additional balancing mechanisms, sparsely gated routers can collapse onto a small subset of experts even when those experts are not globally optimal. Despite this risk, entropy-based objectives remain attractive because they provide differentiable surrogates for efficient end-to-end optimization of selection systems.

Existing approaches handle collapse differently across settings. In sparse MoEs, auxiliary load-balancing objectives are commonly added to prevent routing from concentrating on a small number of experts \citep{shazeer2017, lepikhin2021, fedus2022}. Class imbalance has led to a different set of solutions. Focal loss downweights well-classified examples based on prediction confidence, while class-balanced loss assigns weights according to effective class size \citep{lin2017, cui2019}. Uncertainty can also guide which samples are selected for labeling in active learning \citep{gal2017}. Self-supervised learning avoids related collapse behaviors through architecture-specific mechanisms: DINO uses centering and sharpening, while SimSiam relies on stop-gradient \citep{caron2021, chen2021}. BYOL uses separate online target networks, with the target updated as a moving average of the online network \citep{grill2020}. Other methods learn the weighting rule itself. Self-Tuning Networks optimize hyperparameters through a validation objective, and Meta-Weight-Net learns sample weights from held-out meta-data \citep{mackay2019, shu2019}. Although prior work has studied related entropy-collapse phenomena, they do not provide a general characterization of why entropy alone fails to distinguish confidence from informativeness.

Rather than introducing another domain-specific heuristic, we instead subtract a second entropy term computed from the trust weights before they are aggregated into candidate scores. We prove that this residual-entropy objective removes Shannon entropy's inherent reward for concentration alone. At two extremes, $D\leq0$ holds automatically when trust is distributed uniformly across candidates, while $D\to0$ as the distribution approaches a one-hot limit at any candidate. Between these extremes, our main results give checkable conditions for $D\leq0$. The first is a general pairwise condition that requires the candidate ordering by trust to agree with the ordering by informativeness. A tighter margin condition determines when the leading candidate is sufficiently separated from its competitors. This verifiable confidence margin can then be used as a graded signal for downstream reweighting. Our goal is not to replace the engineering practices that make individual systems work, but to explain why entropy-only selection fails, provide a minimal structural correction, and identify when the resulting candidate ordering can be safely trusted.

Our contributions are fourfold. (1) \emph{Problem.} We formally characterize the arbitrarily placed degenerate minimizers inherent in entropy-only selection objectives (Proposition~1). (2) \emph{Fix.} We derive the exact identity $D=-\mathrm{KL}(p_A\Vert p_\beta)-\Delta$ (Proposition~2), establishing boundary guarantees under uniform and one-hot trust profiles (Corollaries~1--2) and two checkable conditions for safe informativeness-based ordering in the intermediate regime (Corollaries~3--4). (3) \emph{Validation.} We validate the theory across controlled landscapes and independently trained real-world systems, including a direct test of Proposition~2 along a non-stationary optimization trajectory where the conditions of Corollaries~3--4 do not always hold. (4) \emph{Payoff.} In a gradient-trained MoE router, we empirically determine when informativeness-based ordering improves downstream performance: experts must be non-interchangeable, and routing decisions must be used directly rather than averaged. Beyond MoE, the resulting confidence margin supports reweighting that matches or exceeds fixed-strength baselines in a class-imbalance task and reduces MSE by approximately $19$--$20\%$ in independently retrained downstream video prediction, with out-of-sample transfer to a related species.

\section{Related Work}
\label{sec:related}

\textbf{Entropy-driven collapse and self-referential objectives.} Collapse is a recurring problem in objectives that learn from their own predictions. DINO prevents constant outputs through centering and sharpening of the teacher distribution \citep{caron2021}. Self-Classifier takes a different approach, adding a uniform prior to its cross-entropy objective and proving that degenerate assignments are excluded from the set of optimal solutions \citep{amrani2022}. Preventing complete collapse, however, does not necessarily prevent partial collapse. \cite{govindarajan2024} identify partial prototype collapse in the DINO family, where redundant prototypes can persist even when the marginal distribution satisfies the prescribed prior. These results concern particular forms of representation or prototype collapse. Our setting starts one step earlier: minimizing entropy itself can favor an arbitrary concentrated solution because concentration alone does not identify which candidate is informative. Propositions~1--2 characterize this degeneracy directly.
\smallskip

\noindent \textbf{Differentiable selection, routing, and validation-based weighting.} Similar concentration problems appear when a differentiable objective controls a discrete choice. Sparse MoE routers use auxiliary load-balancing objectives to keep traffic from concentrating on a small subset of experts \citep{shazeer2017, lepikhin2021, fedus2022}, while DARTS can increasingly favor skip connections during architecture search \citep{zela2020}. Both motivate corrections tied to the failure mode of the particular system. Here, the correction instead follows from the entropy objective itself and is tested directly on a trained MoE router. A separate literature learns weights or hyperparameters from held-out data rather than from entropy. Self-Tuning Networks optimize hyperparameters through a bilevel training--validation objective \citep{mackay2019}; \cite{Ren2018} derive example weights from their effect on a clean validation loss; and Meta-Weight-Net learns an explicit weighting function using unbiased meta-data \citep{shu2019}. Our downstream weight comes directly from the residual-entropy margin, without a separate validation objective.

\section{Problem Setup and the Collapse Phenomenon}
\label{sec:setup}

We study $T$ candidate groups (e.g., time windows, experts, or hyperparameter bins), each with an outer trust weight $\beta_t>0$ set by an outer optimization loop and an induced candidate score $v_t(\beta_t)\geq0$ held fixed with respect to that loop, giving the induced candidate distribution $p_{A,t}=v_t/\sum_{t'}v_{t'}$ and the outer profile's own distribution $p_{\beta,t}=\beta_t/\sum_{t'}\beta_{t'}$.

\smallskip

\noindent\textbf{Proposition 1 (arbitrarily-placed degenerate minimizers).} \textit{Let $\beta_t$ be a trust profile with induced non-negative scores $v_t(\beta_t)$ that vanish if and only if $\beta_t \to 0$. For any candidate $t_0$, a sequence driving $v_{t_0}$ bounded away from zero while all other $v_t \to 0$ forces $p_A \to \text{one-hot}(t_0)$ and $H(p_A) \to 0$. Because the same minimum entropy can be approached by concentrating on any candidate, minimizing $H(p_A)$ alone provides no preference for a genuinely informative candidate over an arbitrary one.}

\noindent The proof is given in Appendix~\ref{app:proofs}.

\noindent Thus, entropy minimization rewards confidence without identifying informativeness.

\section{The Fix: A Residual-Entropy Objective}
\label{sec:fix}

Define the outer decision variable's own normalized distribution, $p_{\beta,t} = \beta_t/\sum_{t'}\beta_{t'}$, and replace $L_1$ ($H(p_A)$) with the residual objective
\[
L_2(\gamma,\mu) = H(p_A) - H(p_\beta) =: D.
\] 
We still minimize this objective, so lower (more negative) $D$ is what the outer loop favors; the propositions below track whether $D\leq 0$, i.e., at least as good as the untrained baseline.

\smallskip

\noindent\textbf{Proposition 2 (exact characterization of $D$'s sign).} \textit{For any outer trust profile $\beta_t>0$ with induced $p_\beta=\beta_t/\sum_{t'}\beta_{t'}$, and any $p_A$ induced as in Section~\ref{sec:setup},
\[
D \;=\; H(p_A)-H(p_\beta) \;=\; -\,\mathrm{KL}(p_A\Vert p_\beta) \;-\; \Delta, \qquad \Delta \;:=\; \mathbb{E}_{p_A}[\log\beta_t] - \mathbb{E}_{p_\beta}[\log\beta_t],
\]
exactly, with no approximation or assumption on the functional form of $v_t$. Consequently, since $\mathrm{KL}(p_A\Vert p_\beta)\geq0$ always, $D\leq0 \iff \mathrm{KL}(p_A\Vert p_\beta)\geq-\Delta$. In particular, $\Delta\geq0\Rightarrow D\leq0$.}

\begingroup
\setlength{\abovedisplayskip}{3pt plus 1pt minus 1pt}
\setlength{\belowdisplayskip}{3pt plus 1pt minus 1pt}
\setlength{\abovedisplayshortskip}{2pt plus 1pt minus 1pt}
\setlength{\belowdisplayshortskip}{2pt plus 1pt minus 1pt}
\emph{Proof.} By definition,
\[
\mathrm{KL}(p_A\Vert p_\beta) = \sum_t p_{A,t}\log\frac{p_{A,t}}{p_{\beta,t}} = -H(p_A) - \sum_t p_{A,t}\log p_{\beta,t} = -H(p_A) + H^\times(p_A,p_\beta),
\]
so $H(p_A) = H^\times(p_A,p_\beta) - \mathrm{KL}(p_A\Vert p_\beta)$ exactly, and
\[
D = H(p_A)-H(p_\beta) = \big[H^\times(p_A,p_\beta) - H(p_\beta)\big] - \mathrm{KL}(p_A\Vert p_\beta) = -\sum_t(p_{A,t}-p_{\beta,t})\log p_{\beta,t} - \mathrm{KL}(p_A\Vert p_\beta).
\]
Writing $\log p_{\beta,t}=\log\beta_t-\log B$ (with $B=\sum_{t'}\beta_{t'}$) and using $\sum_t(p_{A,t}-p_{\beta,t})=0$, the constant $-\log B$ term cancels:
\[
-\sum_t(p_{A,t}-p_{\beta,t})\log p_{\beta,t} = -\sum_t(p_{A,t}-p_{\beta,t})\log\beta_t = -\Big(\mathbb{E}_{p_A}[\log\beta_t]-\mathbb{E}_{p_\beta}[\log\beta_t]\Big) =: -\Delta.
\]
Hence $D=-\Delta-\mathrm{KL}(p_A\Vert p_\beta)$, $\Delta=\mathbb{E}_{p_A}[\log\beta_t]-\mathbb{E}_{p_\beta}[\log\beta_t]$. Finally, for any two distributions over the same $T$ outcomes, $\mathrm{KL}(p_A\Vert p_\beta)\geq0$, with equality if and only if $p_A=p_\beta$ (Gibbs' inequality). Combined with the identity above, $D\leq0\iff\mathrm{KL}(p_A\Vert p_\beta)\geq-\Delta$, and in particular $\Delta\geq0\Rightarrow D\leq0$, since then $-\Delta\leq0\leq\mathrm{KL}(p_A\Vert p_\beta)$. $\blacksquare$
\endgroup

The condition $\Delta\geq0$ is sufficient but not necessary. When $\Delta<0$, a sufficiently large $\mathrm{KL}(p_A\Vert p_\beta)$ term can still give $D\leq0$. We observe this case on a non-stationary trajectory in Appendix~\ref{app:downstream}.

Two corollaries recover unconditional safety at both extremes of the trust profile's range, without needing $\Delta\geq0$ or any data-dependent assumption at all:

\smallskip

\noindent\textbf{Corollary 1 ($\beta$ driven one-hot at an arbitrary $t_0$).} \textit{If $\beta_{t_0}$ stays bounded below by a fixed positive constant while $\beta_t\to0$ for every $t\neq t_0$, then
\[
p_{\beta,t_0} = \frac{\beta_{t_0}}{\sum_{t'}\beta_{t'}} \to 1, \qquad p_{\beta,t}\to 0 \ (t\neq t_0),
\]
so $p_\beta$ also converges to the one-hot distribution at $t_0$, purely from the shape of the profile, independent of which $t_0$ was chosen. By the same one-hot entropy argument as Proposition~1, $H(p_\beta)\to0$. Combined with $H(p_A)\to0$ from Proposition~1 (which holds for this same $t_0$, by the identical argument), $D=H(p_A)-H(p_\beta)\to0-0=0$ along this sequence, for every choice of $t_0$.} Thus any one-hot limit gives $D \to 0$, regardless of which candidate is selected. This is a tie with the untrained baseline, not an improvement over it.

\smallskip

\noindent\textbf{Corollary 2 ($\beta$ exactly uniform).} \textit{Suppose $\beta_t$ is exactly uniform across all $T$ groups. Then $\log \beta_t$ is the same constant for every $t$, so both $\mathbb{E}_{p_A}[\log \beta_t]$ and $\mathbb{E}_{p_\beta}[\log \beta_t]$ equal that constant for any $p_A$ and $p_\beta$, giving $\Delta = 0$ identically and $D = -\mathrm{KL}(p_A \parallel p_\beta) \le 0$.} Equivalently (uniform $\beta_t$ also gives $p_{\beta,t}=1/T$, so $H(p_\beta)=\log T$, the maximum entropy over $T$ outcomes, while $H(p_A)\leq\log T$ always by the same Jensen's inequality argument applied to $p_A$ directly): $D=H(p_A)-\log T\leq0$, the same conclusion via the more familiar entropy-ceiling framing.

Both corollaries rely strictly on $\beta_t$'s shape being fully known, either concentrated (Corollary~1) or uniform (Corollary~2). In general, however, $\beta_t$ varies dynamically with $v_t$, allowing $\Delta$ to take either sign. The two corollaries below address this general setting by providing checkable, data-dependent sufficient conditions for $\Delta \ge 0$, which by Proposition~2 guarantees $D \le 0$.

\smallskip

\noindent\textbf{Corollary 3 (pairwise sufficient condition for $\Delta\geq0$).} \textit{For any profile $\beta_t$ with induced $v_t$, if for every pair of groups $t,t'$ with $\beta_t\geq\beta_{t'}$, $v_t/\beta_t\geq v_{t'}/\beta_{t'}$, then $\Delta\geq0$, and hence, by Proposition~2, $D\leq0$.}

\smallskip

\noindent\emph{Proof.} Writing $V=\sum_t v_t$ and $B=\sum_t\beta_t$, the condition $\Delta\geq0$, i.e.\ $\mathbb{E}_{p_A}[\log\beta]\geq\mathbb{E}_{p_\beta}[\log\beta]$, becomes, after multiplying through by $VB>0$ and expanding both sides as double sums over $t,t'$,
\[
\sum_{t,t'} v_t\beta_{t'}\log\beta_t \;\geq\; \sum_{t,t'} v_{t'}\beta_t\log\beta_t.
\]
Relabeling the dummy indices $t\leftrightarrow t'$ in the right-hand side and combining, this is equivalent to $\sum_{t,t'}v_t\beta_{t'}(\log\beta_t-\log\beta_{t'})\geq 0$, which (pairing each $(t,t')$ term with its $(t',t)$ counterpart) equals
\[
S := \sum_{t<t'}\left(v_t\beta_{t'}-v_{t'}\beta_t\right)\left(\log\beta_t-\log\beta_{t'}\right) \;\geq\; 0, \qquad\text{where}\quad S = VB\cdot\Delta.
\]
For any pair $t,t'$ with, say, $\beta_t\geq\beta_{t'}$ (so $\log\beta_t-\log\beta_{t'}\geq 0$), the factor $(v_t\beta_{t'}-v_{t'}\beta_t)$ has the same sign as $(v_t/\beta_t - v_{t'}/\beta_{t'})$ (dividing by $\beta_t\beta_{t'}>0$). If $v_t/\beta_t\geq v_{t'}/\beta_{t'}$ whenever $\beta_t\geq\beta_{t'}$, for every pair, then every term of $S$ is a product of two non-negative factors, so $S\geq0$, hence $\Delta\geq0$, hence, by Proposition~2, $D\leq0$. $\blacksquare$ (Verified numerically against $0$ violations across $50{,}000$ random landscapes, Appendix~\ref{app:proofs}.)

\smallskip

\noindent\textbf{Corollary 4 (a tighter bound on $\Delta$, empirically motivated).} \textit{Let $w=\arg\max_t\beta_t=\arg\max_t(v_t/\beta_t)$ (both hold automatically once the outer loop has settled on a single, genuinely dominant candidate), $O=\{1,\dots,T\}\setminus\{w\}$, $\Delta_{\min}=\min_{t\in O}(v_w/\beta_w-v_t/\beta_t)\geq0$ ($w$'s smallest ratio margin over any competitor), and $M_O=\sum_{t\in O}\beta_t$. If
\[
\Delta_{\min}\cdot\beta_w\sum_{t\in O}\beta_t(\log\beta_w-\log\beta_t) \;\geq\; M_O^2\cdot\sigma_O(\log\beta)\cdot\sigma_O(v/\beta),
\]
where $\sigma_O$ is the $\beta$-weighted standard deviation over $O$, then $D\leq0$.}

\smallskip

\noindent\emph{Proof.} Recall from Corollary~3's proof that $\Delta\geq0\iff S\geq0$ where $S:=VB\cdot\Delta=\sum_{t<t'}(v_t\beta_{t'}-v_{t'}\beta_t)(\log\beta_t-\log\beta_{t'})$. Split $S=S_w+S_O$ (pairs involving $w$; pairs within $O$ only). Using the standard identity $\sum_{t<t',\,t,t'\in O}\beta_t\beta_{t'}(x_t-x_{t'})(y_t-y_{t'})=M_O^2\,\mathrm{Cov}_O(x,y)$ for weights $\beta_t$ summing to $M_O$ (verified directly, both symbolically and against $20$ random landscapes, zero mismatches), with $x_t=\log\beta_t$ and $y_t=v_t/\beta_t$, $S_O=M_O^2\cdot\mathrm{Cov}_O(\log\beta,v/\beta)$, the $\beta$-weighted covariance between $\log\beta_t$ and $v_t/\beta_t$ over $t\in O$. Since $\beta_w\geq\beta_t$ and $v_w/\beta_w\geq v_t/\beta_t$ for every $t\in O$, each term of $S_w$ is a product of two non-negative factors. Replacing each margin with its minimum $\Delta_{\min}$ gives $S_w\geq\Delta_{\min}\cdot\beta_w\sum_{t\in O}\beta_t(\log\beta_w-\log\beta_t)$. By Cauchy--Schwarz, $|\mathrm{Cov}_O(\log\beta,v/\beta)|\leq\sigma_O(\log\beta)\cdot\sigma_O(v/\beta)$, so $S_O\geq-M_O^2\sigma_O(\log\beta)\sigma_O(v/\beta)$. Combining, $S=S_w+S_O\geq\Delta_{\min}\beta_w\sum_{t\in O}\beta_t(\log\beta_w-\log\beta_t)-M_O^2\sigma_O(\log\beta)\sigma_O(v/\beta)\geq0$ under the stated condition, so $\Delta\geq0$, hence $D\leq0$. $\blacksquare$ (Verified numerically: holds on $12.94\%$ of $20{,}000$ random landscapes with zero violations, and at all $60/60$ examined converged points from studies (i) and (iv), Appendix~\ref{app:proofs}.)

\begin{figure}[h]
\centering
\includegraphics[width=\linewidth]{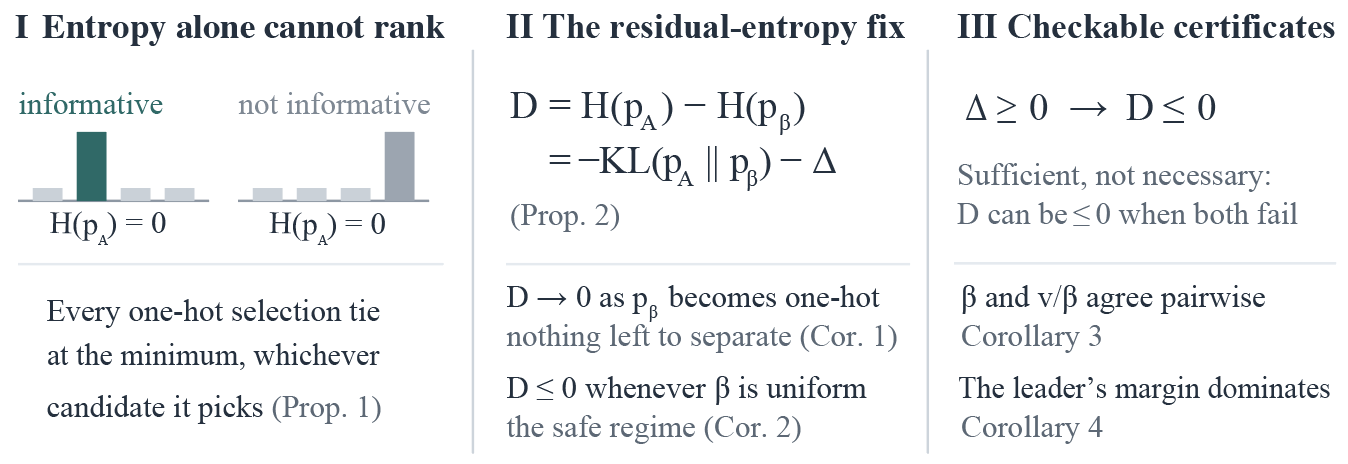}
\caption{From entropy degeneracy to residual-entropy correction. Propositions~1--2 characterize the problem and correction and Corollaries~1--4 provide boundary results and sufficient sign certificates.}
\label{fig:summary}
\end{figure}
\FloatBarrier

As summarized in Figure~\ref{fig:summary}, the theoretical framework establishes the degeneracy of minimizing entropy alone, the residual entropy correction, and their associated corollaries. Subsequent Sections~5 and~6 numerically verify these theories and evaluate whether Corollary~2 for uniform classes as well as Corollaries~3 and~4 for correct ordering can serve as diagnostics and enhance downstream gains across mixture of experts routing, class imbalance, and a separately developed prediction pipeline. Our findings demonstrate that correct ordering improves downstream performance only when the selection is directly consumed by a downstream action and the candidates are not interchangeable.

\section{Empirical Validation}
\label{sec:validation}

Every study below instantiates the same computational pipeline (Algorithm~1), built by default on the same rational scoring form $v_t=\sum_k c_{t,k}\beta_t/(\beta_t+a_{t,k})$ and outer profile $\beta_t(\theta)=\beta_{\max}\exp[-\gamma(t-\mu)^2]$ ($\theta=(\gamma,\mu)$). Studies that instead use a different functional form or profile shape note this explicitly where they are introduced.

\begin{algboxhere}{1}{Generic outer-loop protocol shared by every study in Section~\ref{sec:validation}. \label{alg:generic}}
\textbf{Require:} $T$ candidate groups; outer-profile parameters $\theta$; per-candidate score function $v_t(\beta_t)$; objective $L\in\{L_1,L_2\}$, $L_1=H(p_A)$, $L_2=H(p_A)-H(p_\beta)$\par
\algln Initialize $\theta$ at the study's $\mu$-initialization\par
\algln \textbf{while} not converged \textbf{do}\par
\hspace{1.5em}\algln $\beta_t \gets \beta_t(\theta)$ for every group $t$ \hfill\textit{// outer trust weight}\par
\hspace{1.5em}\algln $v_t \gets v_t(\beta_t)$ for every group $t$ \hfill\textit{// induced score; fixed w.r.t.\ $\theta$ unless non-stationary}\par
\hspace{1.5em}\algln $p_{A,t} \gets v_t/\sum_{t'}v_{t'}$, \quad $p_{\beta,t}\gets\beta_t/\sum_{t'}\beta_{t'}$\par
\hspace{1.5em}\algln $D \gets H(p_A)-H(p_\beta)$ \hfill\textit{// Proposition~2's identity, evaluated exactly at every step}\par
\hspace{1.5em}\algln $\theta \gets \theta - \eta\,\nabla_\theta L$ \hfill\textit{// Adam; finite-difference or true backprop depending on the study}\par
\algln \textbf{end while}\par
\algln \textbf{return} $\theta^\ast,\beta^\ast,p_A^\ast,p_\beta^\ast,D^\ast$ \hfill\textit{// feeds diagnostic: ratio, selection accuracy, margin $m$}
\end{algboxhere}

\subsection{Core Mechanism Validation}

Studies (i)--(iii) share one synthetic, architecture-decoupled landscape ($T{=}5$, $K{=}50$, one informative group, 30 seeds $\times$ 3 $\mu$-initializations; full setup in Appendix~\ref{app:evidence}). The statistic that matters throughout Table~\ref{tab:master} for these three studies is \emph{ratio}: mean converged objective value when optimization starts at the informative candidate, divided by the mean value when it starts elsewhere.

\smallskip

\noindent\textbf{(i) Initialization ablation.} We compare converged objectives from informative and elsewhere initializations (Table~\ref{tab:master}). Entropy alone reaches nearly equal values, consistent with Proposition~1's arbitrary-placement degeneracy. Residual entropy reaches a lower value from informative initialization, allowing comparison across initializations to favor that candidate.

\smallskip

\noindent\textbf{(ii) Functional-form ablation.} We test sigmoid and logarithmic scores and softmax-based gating. Residual entropy favors informative initialization under rational and logarithmic scoring. Sigmoid and softmax gating yield equal mean objectives across initialization groups. Proposition~2's identity applies across these forms, while initialization dependence varies (Appendix~\ref{app:evidence}).

\noindent\textbf{(iii) Alternative-objective ablation.} We test a concentration penalty, a tighter concentration bound, a penalty pulling $p_\beta$ toward uniform, and $H(p_A)-\mathrm{KL}(p_A\Vert p_\beta)$. Standard penalties produce some objective separation. The KL variant retains the initialization-associated window in all 90 runs. At common one-hot collapse, its KL term vanishes, leaving the entropy-only objective unchanged. Residual entropy also tends to zero there but attains negative values away from collapse, including $-0.3641$ at informative initialization. These comparisons distinguish objective separation from collapse behavior. They do not establish uniqueness among all possible corrections (Appendix~\ref{app:evidence}).

\begin{table}[h]
\caption{Seven studies: questions and quantitative evidence. Objective pairs are informative / elsewhere initialization means.}
\label{tab:master}
\centering
\tablecaptionspace
\footnotesize
\setlength{\tabcolsep}{3pt}
\renewcommand{\arraystretch}{0.92}
\begin{tabularx}{\linewidth}{@{}p{0.22\linewidth}p{0.29\linewidth}X@{}}
\toprule
Study / target & Entropy alone / alternatives & Residual entropy \\
\midrule
(i) Initialization: objective separation & 0.0047 / 0.0050; ratio 0.93 & $-0.3641$ / $-0.0134$; ratio 27.11 \\
(ii) Functional form: scoring / normalization & Ratios 0.64--0.93 & Rational / log: 27.11 / 23.67; sigmoid / softmax: 1.00 / 1.00 \\
(iii) Alternatives: separation / collapse & Penalty ratios: 0.62--0.88; KL variant: 90/90 retain initial window & $-0.3641$ / $-0.0134$; lower objective at informative initialization \\
(iv) Class imbalance: identification & Selection rate 0.775 & 0.938 ($1.21\times$; $p=0.003$) \\
(v) Full-batch MoE: participation boundary & 5/6 lower-bounded at floor$>0$; no transition & Threshold 0.07--0.10; 0/6 below, 4/6 above \\
(vi) Streaming MoE: boundary robustness & 5/6 lower-bounded at floor$>0$; no transition & Threshold 0.062--0.070; 0/6 below, 4/6 above \\
(vii) Real pipeline: initialization consistency & W1, W5, W5 & W3, W3, W3 \\
\bottomrule
\end{tabularx}
\end{table}

\FloatBarrier

\subsection{Diagnostic Capability Evidence}

Study (iv) fixes the inner quantities $\{c_{t,k},a_{t,k}\}$, matching the stationary theory. In studies (v)--(vi), they evolve with training. Proposition~2's identity remains applicable, while its sufficient conditions must be checked along the trajectory.

\smallskip

\noindent\textbf{(iv) Real-data class-imbalance selection (stationary).} \emph{Setup:} a real-data digits task ($T{=}5$) at two scarcity levels (8\% and 2\%). One class is made scarce by reducing its training examples to the stated fraction of the original count. We ask whether the residual-entropy formulation identifies the scarce class more reliably than $H(p_A)$ alone, and whether acting on that identification through oversampling improves downstream class-imbalance performance.

\smallskip

\noindent\emph{Diagnostic method:} we construct the algorithm as described in Appendix~\ref{app:digits}. A one-vs-rest logistic-regression probe contributes the per-example confusion signal used to score class imbalance. At convergence, the class with the largest $\beta$ value is designated as the scarce class for each seed, using the majority vote across $T$ independent $\mu$-initializations in Algorithm~2. To translate this identification into downstream oversampling, we compute the margin-derived confidence $m=(r_{t^\ast}-r_{t'})/(r_{t^\ast}+r_{t'}+\epsilon)\in(-1,1)$, which measures how much more strongly the identified candidate is favored than its closest competitor, where $r_t=v_t/\beta_t$ is Corollary~3's compared quantity, $t^\ast=\arg\max_t\beta_t$ is the identified class, and $t'$ is its strongest competitor according to $r_t$. We then set $\mathrm{confidence}=\mathrm{clip}(m/s,0,1)$ for a scale $s$, giving a downstream oversampling factor of $1+4\times\mathrm{confidence}$. Algorithm~2 in Appendix~\ref{app:digits} provides the full computation.

\smallskip

\noindent\emph{Diagnostic result:} at 8\% scarcity (minority class~3, $n=80$), selection improves from 0.775 to 0.938 ($1.21\times$; $z=2.93$, $p=0.003$). Under entropy alone, four initializations ($\mu=0,1,3,4$) retain their initial class in all 80 seeds. Only $\mu=2$ moves, reaching class~3 in 62/80 seeds and deciding the majority vote. Residual entropy breaks this retention: $\mu=4$ reaches class~3 in every seed, while $\mu=2$ also always moves. Its five misidentifications reflect a vote split for class~1 rather than fixed boundary retention. At convergence, $D\leq0$ in 396/400 runs (mean $D=-0.20$). The remaining four satisfy $D\leq1.4\times10^{-4}$.

\smallskip

\noindent\emph{Diagnostic payoff:} we test whether using this identification in downstream retraining improves minority-class recall by comparing a fixed $5\times$ minority-class oversampling factor with confidence-scaled oversampling ($1\times$--$5\times$, $s{=}0.25$). Table~\ref{tab:confreweight} reports both scarcity levels, with the full method given in Appendix~\ref{app:confidence}. Confidence-scaled reweighting never underperforms uniform training in any evaluated seed. Its advantage over the fixed factor is not statistically significant at 8\% scarcity ($p{=}0.159$), because only 5 of 80 seeds are misidentified, leaving little room for the methods to diverge. Nevertheless, it eliminates all harmful cases produced by the fixed factor ($0/80$ versus $2/80$). At 2\% scarcity, where more seeds are misidentified, confidence scaling significantly outperforms both the fixed factor ($p{=}0.023$) and uniform training ($p{=}0.012$). Thus, acting on informativeness-based selection through confidence-scaled oversampling converts the diagnostic margin into downstream value while limiting the harm caused by incorrect selections.

\begin{table}[h]
\caption{Confidence-modulated reweighting at both scarcity levels: mean held-out minority-class recall after retraining with each oversampling scheme. ``Hurt'' counts seeds below uniform training; the final column compares confidence-scaled with fixed $5\times$ weighting. The same $s{=}0.25$ is used.}
\label{tab:confreweight}
\centering
\tablecaptionspace
\small
\begin{tabular}{@{}lrrrrr@{}}
\toprule
Setting & Uniform & Fixed $5\times$ & Conf.-scaled & Hurt (fixed / conf.) & $p$ (conf.\ vs.\ fixed) \\
\midrule
Digits $T{=}5$, 8\% ($n{=}80$) & 0.8466 & 0.8727 & 0.8732 & 2/80 \, / \, 0/80 & 0.159 \\
Digits $T{=}5$, 2\% ($n{=}15$) & 0.6776 & 0.6909 & 0.7018 & 5/15 \, / \, 0/15 & 0.023 \\
\bottomrule
\end{tabular}
\end{table}

\smallskip

\noindent\textbf{(v)--(vi) Mixture-of-experts routing collapse: diagnostic extension to a coupled, non-stationary bilevel system.} \emph{Setup:} a gradient-trained MoE consisting of $T{=}6$ softmax-classifier experts on scikit-learn's digits dataset, under two data-exposure regimes: full-batch training at every outer step (study (v), Appendix~\ref{app:moe}) and incremental streaming through 15 sequential chunks (study (vi), Appendix~\ref{app:bandit}). One expert receives a measured warm-start advantage. Here, $\{c_{e,k}\}$ evolves during training, so unlike in study (iv), Proposition~2's checkable corollaries need not hold throughout. Instead, we diagnose which regime the trajectory reaches.

\smallskip

\noindent\emph{Diagnostic method:} $c_{e,k}$ is coupled to $\beta_e$ through a real-time rich-get-richer training loop. An expert at the lower bound trains more slowly and therefore achieves lower measured accuracy at the next step. Thus, $c_{e,k}$ is not detached from $\beta_e$, as assumed in Proposition~1. Corollary~2 ($\gamma{=}0\Rightarrow D\leq0$ for any $c,a$) still applies under this coupling. We sweep \emph{floor}, the lower bound in
\[
\mathrm{lr\_mult}_e=\operatorname{clip}\left(T\beta_e/\bar\beta,\,\mathrm{floor},5.0\right),
\]
from $0$ to $1.0$ and count the experts that remain at the lower bound at convergence. We use this count rather than the raw sign of $D$, since $D$ can become transiently positive during coupled training. At $\mathrm{floor}=0$, a lower-bounded expert is literally starved. At $\mathrm{floor}>0$, it continues training at the floor-capped rate. All tables use the fair protocol described in Appendix~\ref{app:moe-fair}.

\smallskip

\noindent\emph{Diagnostic result:} under $H(p_A)$ alone, the floor is diagnostically inert. 5/6 experts remain lower-bounded at every $\mathrm{floor}>0$, and 4/6 are literally starved at $\mathrm{floor}=0$, with no transition anywhere from $0$ to $1.0$. Under the residual objective, a narrow threshold separates two regimes: approximately $0.07$--$0.10$ in study (v) and $0.062$--$0.070$ in study (vi). Below it, $\gamma$ reaches the regime characterized by Corollary~2 ($\gamma\approx0$), and lower-bounding is eliminated ($0/6$ experts). Above it, $\gamma$ remains peaked ($\gamma_f\approx4.1$), and $4/6$ experts remain lower-bounded (Table~\ref{tab:master}; Appendices~\ref{app:moe-fair} and~\ref{app:bandit-fair}).

\smallskip

\noindent\emph{Diagnostic payoff:} the threshold shifts with the data distribution, making floor a control parameter. Across seven noise levels ($n{=}15$ seeds per level), it falls from approximately $0.10$--$0.12$ under clean data to $0.02$--$0.05$ under moderate noise and vanishes above approximately $1.5$--$2.0$. Which side of the threshold achieves higher accuracy also depends on the noise level, so the diagnostic must be evaluated per deployment (Table~\ref{tab:floor-threshold-noise}; Appendix~\ref{app:moe-fair}).

\begin{table}[h]
\caption{Floor threshold and accuracy vs.\ task noise (study (v), residual, fair protocol, $n{=}15$ seeds/noise level). ``Threshold'' separates clamped ($\gamma_f\approx0$) from peaked ($\gamma_f\approx4$); ``Winner'' has higher test acc.\ (Appendix~\ref{app:moe-fair}).}
\label{tab:floor-threshold-noise}
\centering
\tablecaptionspace
\footnotesize
\setlength{\tabcolsep}{3pt}
\renewcommand{\arraystretch}{0.78}
\begin{tabular}{@{}lrrrl@{}}
\toprule
Noise & Threshold & Clamped acc. & Peaked acc. & Winner \\
\midrule
0.0 & 0.10--0.12 & 0.960 & 0.964 & peaked ($+0.3$pt) \\
0.5 & 0.09--0.10 & 0.924 & 0.918 & clamped ($+0.6$pt) \\
1.0 & 0.07--0.08 & 0.803 & 0.783 & clamped ($+2.0$pt) \\
1.5 & 0.02--0.05 & 0.660 & 0.645 & clamped ($+1.5$pt) \\
$\ge2.0$ & vanished & -- & $0.37$--$0.54$ & n/a (single regime) \\
\bottomrule
\end{tabular}
\end{table}

Safety (Corollary~2) and ordering (Corollaries~3--4) provide independent guarantees. Having established the value of safety above, we next ask when guaranteeing correct ordering is also useful. We test candidate interchangeability in the same MoE setup by removing a fraction of input pixels, denoted \emph{gap}, from the experts without the warm-start advantage. As gap increases from $0$ to $0.5$, the benefit of correct ordering increases from $-0.003$ to $+0.141$ ($\mathrm{slope}=+0.280$, $r=0.94$). The payoff from correct ordering therefore grows as candidates become less interchangeable (Appendix~\ref{app:fungibility}). The safety result is unchanged.

\FloatBarrier
\section{Real-Pipeline Validation and Downstream Application (Study vii)}
\label{sec:case-study}

\begin{figure}[h]
\centering
\includegraphics[width=0.744\linewidth]{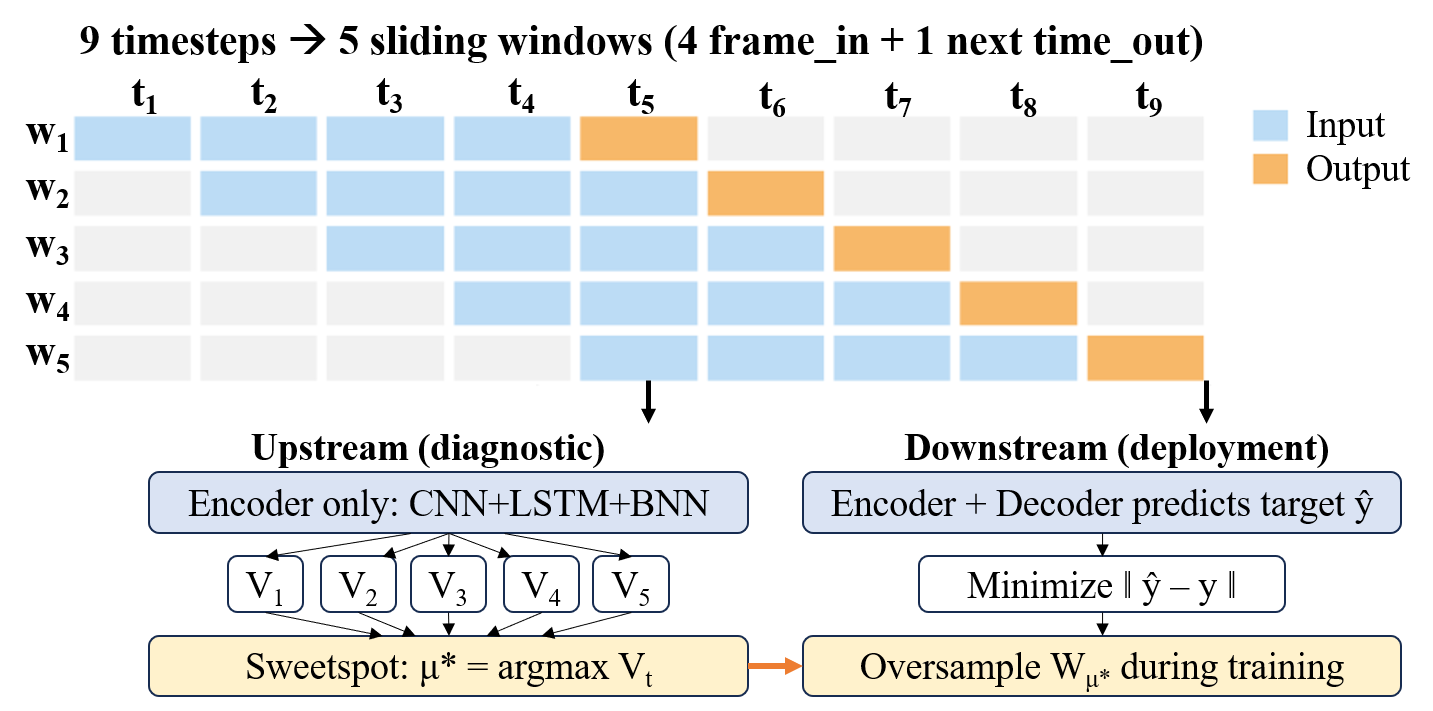}
\caption{Real-pipeline architecture. An encoder-only upstream model identifies a sweet-spot window among five candidates. That window is then oversampled when training a separate encoder-decoder predictor. The exact upstream score is derived in Appendix~\ref{app:downstream-derivation}.}
\label{fig:pipeline}
\end{figure}

\subsection{Reliability Ablation}
\label{sec:real-ablation}

We re-train the production pipeline's upstream uncertainty model on its own real data, swapping only the outer-loop objective ($L_1$ vs.\ $L_2$), across three $\mu$-initializations. $H(p_A)$ alone lets $\gamma$ diverge and the selected window is not reproducible. The residual objective keeps $\gamma$ bounded and all three converge to the same window (Table~\ref{tab:master}; Appendix~\ref{app:downstream}). Unlike study (iv), this system is \textbf{non-stationary}: $v_t$ evolves live with training, so neither Corollary~3 nor Corollary~4 need hold at convergence. When checked directly, neither does (Appendix~\ref{app:downstream}). Despite this, Proposition~2's general identity remains directly evaluable regardless of stationarity, and holds at all three converged checkpoints ($D<0$ each), giving a real safety guarantee precisely where the checkable corollaries alone would not.

\subsection{Downstream Application to a Separately-Developed Predictor}

To test whether the mechanism has consequences beyond the optimization trajectory, we evaluate it downstream. The upstream model selects a single sweet-spot window $\mu^\ast$ among five candidate windows of a growth-curve video (construction and derivation in Appendix~\ref{app:downstream-derivation}). We then oversample that window $5\times$ when training a separately developed sliding-window video predictor (Figure~\ref{fig:pipeline}) with identical architecture across arms.

We compare five arms differing only in which window (if any) is oversampled: \texttt{uniform} (no emphasis, the ``before'' condition); \texttt{random} and \texttt{raw\_var} (an uninformative control and a model-free pixel-variance heuristic); \texttt{mc\_var} (a standard MC-dropout/BALD-style predictive-variance baseline computed directly on the downstream predictor, without the upstream model); and \texttt{sweetspot} ($\mu^\ast$ itself). Each arm is evaluated not only at its own target window but at all five, on held-out synthetic data and two fully independent real videos, pooling the resulting 13 (window, dataset) points into one mean$\pm$SD MSE per arm.

\begin{table}[!ht]
\caption{Bacillus, all five windows pooled (13 evaluation points: 5 held-out synthetic + 5 video1 + 3 video2 windows; lower MSE is better).}
\label{tab:allwindow}
\centering
\tablecaptionspace
\footnotesize
\begin{tabular}{@{}lrr@{}}
\toprule
Arm & Mean MSE & SD \\
\midrule
\texttt{sweetspot} ($\mu^\ast$, ours) & \textbf{92.4} & 82.0 \\
\texttt{uniform} (no oversampling) & 104.0 & 82.5 \\
\texttt{mc\_var} (BALD-style) & 106.5 & 91.7 \\
\texttt{random} & 116.1 & 85.3 \\
\texttt{raw\_var} & 139.9 & 115.6 \\
\bottomrule
\end{tabular}
\end{table}

While only the \texttt{sweetspot} method achieves gains over the \texttt{uniform} baseline, all alternatives, including \texttt{mc\_var}, underperform relative to doing nothing even within their own oversampled windows, confirming that the improvement derives specifically from targeting the upstream model's identified window rather than from oversampling alone (Appendix~\ref{app:downstream}). We further test transfer to a different microalgal species (\textit{Tetraselmis}) and a structurally unrelated eukaryote (HeLa). The advantage is smaller on \textit{Tetraselmis} and disappears on the more distant HeLa data, where \texttt{mc\_var} performs better (Appendix~\ref{app:downstream}).

\FloatBarrier
\section{Conclusion}

This paper begins with a simple limitation of Shannon entropy. A low value of $H(p_A)$ indicates that selection is concentrated, but says nothing about whether it has concentrated on an informative candidate. Proposition~1 makes this ambiguity explicit by showing that the same minimum can be approached at any candidate. Subtracting $H(p_\beta)$ removes the reward for concentration alone and gives the exact decomposition $D=-\mathrm{KL}(p_A\Vert p_\beta)-\Delta$. From this identity, the boundary cases provide safety under uniform and one-hot trust, while Corollaries~3--4 give checkable conditions for $D\leq0$ when trust lies between these extremes. 

Across synthetic landscapes, a class-imbalance task, and a gradient-trained MoE router, residual entropy separates informative preference from confidence alone. In the MoE system, it reveals a data-dependent threshold governing routing collapse and accuracy. Controlled experiments further show that correct ordering improves downstream performance when candidates are non-interchangeable and selections are acted upon directly, whereas safety does not require candidate differences.

We finally test whether the selected preference remains useful outside these validation settings. In an independently developed video-prediction pipeline, oversampling the window identified upstream reduces MSE by approximately \(19\)--\(20\%\) compared with uniform training. The improvement persists on a related species, although it narrows, and reverses relative to the predictive-variance baseline on the more domain-distant data. The downstream benefit therefore extends beyond the systems used to validate the theory, but does not transfer without limit.

\subsection*{AI use statement}
Generative AI tools were used to support manuscript editing, literature retrieval, and code development. They were also used to generate synthetic datasets used in experiments (i)--(iii). Research hypotheses, mathematical claims and proofs, and experimental design were developed by the authors. The authors independently reviewed and verified all AI-assisted code, text, and citations, and take responsibility for the complete contents of this work. 

\subsection*{Reproducibility statement}
Complete proofs and numerical verification of the theoretical results are provided in Section~\ref{sec:fix} and Appendix~\ref{app:proofs}. Supplementary materials include standalone code, exact commands, and experimental details for studies (i)--(vi) and the confidence-modulated reweighting experiments. Study (vii) uses a non-public production pipeline; its ablation harness and raw evaluation outputs are included so that the reported tables and figures can be independently checked. Additional reproducibility details for each study appear in the corresponding appendix section.

\bibliographystyle{plainnat}
\bibliography{ICLR2027}

@inproceedings{dai2024,
    title = "{D}eep{S}eek{M}o{E}: Towards Ultimate Expert Specialization in Mixture-of-Experts Language Models",
    author = "Dai, Damai  and
      Deng, Chengqi  and
      Zhao, Chenggang  and
      Xu, R.X.  and
      Gao, Huazuo  and
      Chen, Deli  and
      Li, Jiashi  and
      Zeng, Wangding  and
      Yu, Xingkai  and
      Wu, Y.  and
      Xie, Zhenda  and
      Li, Y.K.  and
      Huang, Panpan  and
      Luo, Fuli  and
      Ruan, Chong  and
      Sui, Zhifang  and
      Liang, Wenfeng",
    booktitle = "ACL",
    year = "2024",
    pages = "1280--1297"
}

@inproceedings{li2024,
	title = {Merge, then compress: {Demystify} efficient {SMoE} with hints from its routing policy},
	booktitle = {ICLR},
	author = {Li, Pingzhi and Zhang, Zhenyu and Yadav, Prateek and Sung, Yi-Lin and Cheng, Yu and Bansal, Mohit and Chen, Tianlong},
	year = {2024}
}

@inproceedings{ren2018,
  title={Learning to Reweight Examples for Robust Deep Learning},
  author={Mengye Ren and Wenyuan Zeng and Bin Yang and Raquel Urtasun},
  booktitle={ICML},
  volume={80},
  pages={4334--4343},
  year={2018}
}

@inproceedings{zela2020,
  title={Understanding and Robustifying Differentiable Architecture Search},
  author={Arber Zela and Thomas Elsken and Tonmoy Saikia and Yassine Marrakchi and Thomas Brox and Frank Hutter},
  booktitle={ICLR},
  year={2020}
}

@inproceedings{govindarajan2024,
  title={On Partial Prototype Collapse in the {DINO} Family of Self-Supervised Methods},
  author={Hariprasath Govindarajan and Per Sid{\'e}n and Jacob Roll and Fredrik Lindsten},
  booktitle={BMVC},
  year={2024}
}

@inproceedings{amrani2022,
  title={Self-Supervised Classification Network},
  author={Elad Amrani and Leonid Karlinsky and Alex Bronstein},
  booktitle={ECCV},
  year={2022},
  pages={116--132}
}

@inproceedings{shu2019,
  title={Meta-Weight-Net: Learning an Explicit Mapping for Sample Weighting},
  author={Jun Shu and Qi Xie and Lixuan Yi and Qian Zhao and Sanping Zhou and Zongben Xu and Deyu Meng},
  booktitle={NeurIPS},
  volume={32},
  pages={1919--1930},
  year={2019}
}

@inproceedings{mackay2019,
  title={Self-Tuning Networks: Bilevel Optimization of Hyperparameters Using Structured Best-Response Functions},
  author={Matthew MacKay and Paul Vicol and Jon Lorraine and David Duvenaud and Roger Grosse},
  booktitle={ICLR},
  year={2019}
}

@inproceedings{grill2020,
  title={Bootstrap Your Own Latent: A New Approach to Self-Supervised Learning},
  author={Jean-Bastien Grill and Florian Strub and Florent Altch{\'e} and Corentin Tallec and Pierre H. Richemond and Elena Buchatskaya and Carl Doersch and Bernardo Avila Pires and Zhaohan Daniel Guo and Mohammad Gheshlaghi Azar and Bilal Piot and Koray Kavukcuoglu and R{\'e}mi Munos and Michal Valko},
  booktitle={NeurIPS},
  volume={33},
  pages={21271--21284},
  year={2020}
}

@inproceedings{chen2021,
  title={Exploring Simple {S}iamese Representation Learning},
  author={Xinlei Chen and Kaiming He},
  booktitle={CVPR},
  pages={15745--15753},
  year={2021}
}

@inproceedings{caron2021,
  title={Emerging Properties in Self-Supervised Vision Transformers},
  author={Mathilde Caron and Hugo Touvron and Ishan Misra and Herv{\'e} J{\'e}gou and Julien Mairal and Piotr Bojanowski and Armand Joulin},
  booktitle={ICCV},
  pages={9650--9660},
  year={2021}
}

@inproceedings{gal2017,
  title={Deep {B}ayesian Active Learning with Image Data},
  author={Yarin Gal and Riashat Islam and Zoubin Ghahramani},
  booktitle={ICML},
  volume={70},
  pages={1183--1192},
  year={2017}
}

@inproceedings{cui2019,
  title={Class-Balanced Loss Based on Effective Number of Samples},
  author={Yin Cui and Menglin Jia and Tsung-Yi Lin and Yang Song and Serge J. Belongie},
  booktitle={CVPR},
  pages={9268--9277},
  year={2019}
}

@article{fedus2022,
  title={Switch Transformers: Scaling to Trillion Parameter Models with Simple and Efficient Sparsity},
  author={William Fedus and Barret Zoph and Noam Shazeer},
  journal={Journal of Machine Learning Research},
  volume={23},
  number={120},
  pages={1--39},
  year={2022}
}

@inproceedings{
lepikhin2021,
title={GShard: Scaling giant models with conditional
computation and automatic sharding},
author={Dmitry Lepikhin and HyoukJoong Lee and Yuanzhong Xu and Dehao Chen and Orhan Firat and Yanping Huang and
Maxim Krikun and Noam Shazeer and Zhifeng Chen},
booktitle={ICLR},
year={2021}
}

@inproceedings{lin2017,
  title={Focal Loss for Dense Object Detection},
  author={Tsung-Yi Lin and Priya Goyal and Ross B. Girshick and Kaiming He and Piotr Doll{\'a}r},
  booktitle={ICCV},
  pages={2980--2988},
  year={2017}
}

@inproceedings{
shazeer2017,
title={Outrageously Large Neural Networks: The Sparsely-Gated Mixture-of-Experts Layer},
author={Noam Shazeer and Azalia Mirhoseini and Krzysztof Maziarz and Andy Davis and Quoc Le and Geoffrey Hinton and Jeff Dean},
booktitle={ICLR},
year={2017}
}

\newpage
\appendix

\section{Proofs and Numerical Verification}
\label{app:proofs}

We first prove Proposition~1, followed by numerical verification of the checkable corollaries. Proposition~2 and Corollaries~1--4 are proved in Section~\ref{sec:fix}.

\smallskip

\noindent\textbf{Proof of Proposition~1.} Fix any target candidate $t_0$. By assumption, there is a sequence of outer-variable settings along which $\beta_{t_0}$ stays bounded below by some fixed $c>0$ while $\beta_t\to0$ for every $t\neq t_0$. By hypothesis this gives $v_t\to0$ for every $t\neq t_0$ along the same sequence, while $v_{t_0}$ stays bounded below by some fixed positive constant throughout. Therefore $p_{A,t_0}=v_{t_0}/\sum_{t'}v_{t'}\to1$ and $p_{A,t}\to0$ for $t\neq t_0$: $p_A$ converges to the one-hot distribution at $t_0$. Since $H(p)\geq0$ with equality if and only if $p$ is one-hot, $H(p_A)\to0$ is the global minimum, attained along this sequence for every choice of $t_0$. Consequently, minimizing $H(p_A)$ alone admits a degenerate family of globally optimal solutions, none of which depends on whether the selected candidate is genuinely informative. $\blacksquare$

\smallskip

\noindent\textbf{Remark (shared noise scale).} When $v_t$ takes the paper's rational scoring form $v_t=\sum_k c_{t,k}\beta_t/(\beta_t+a_{t,k})$ (Section~\ref{sec:setup}, used throughout studies (i)--(vii)) with a shared scale $a_{t,k}=a$ for all $t,k$, writing $C_t:=\sum_k c_{t,k}$ for group $t$'s total informativeness, Corollary~3's condition simplifies to $C_t/C_{t'}\geq\beta_t/\beta_{t'}$ whenever $\beta_t\geq\beta_{t'}$. This condition also matches an empirical pattern observed throughout Sections~\ref{sec:validation}--\ref{sec:case-study} (e.g., Appendix~\ref{app:evidence}): the algorithm converges rather than diverging, consistent with $\beta_t/\beta_{t'}$ growing past what $C_t/C_{t'}$ supports.

\smallskip

\noindent\textbf{Corollary~3.} Corollary~3 provides a sufficient rather than necessary condition for $D \leq 0$, derived purely algebraically in Appendix~A. To verify this derivation numerically across $20{,}000$ random landscapes with parameters $T \in [2,8)$, $K \in [1,10)$, and $\{c_{t,k}, a_{t,k}, \beta_t\}$ drawn independently from uniform distributions (where $v_t = \sum_k \frac{c_{t,k}\beta_t}{\beta_t + a_{t,k}}$), the condition held in 662 cases. Crucially, $D \leq 0$ was satisfied in all 662 instances with zero violations, confirming that the algebraic derivation contains no sign errors or missed edge cases. Because the condition is strictly sufficient, $D \leq 0$ can still hold when the condition fails. Among the 19,338 failing cases, $D$ was still nonpositive 16\% of the time (and positive 84\% of the time), which is consistent with an informative yet incomplete bound. Additionally, under a single shared noise scale $a$, the condition held in 830 out of a further 30,000 random landscapes, again with zero violations. All findings are fully reproducible using \texttt{verify\_proposition2\_corollary3.py} in the supplementary materials.

\smallskip

\noindent\textbf{Corollary~4.} As demonstrated above, Corollary~3's full pairwise condition almost never holds at the points gradient descent actually reaches. This empirical limitation motivated Corollary~4 in Section~\ref{sec:fix}, which directly targets conditions that realistic optimization trajectories satisfy. On the same 20,000 random landscapes generated from the paper's $\beta_t(\gamma, \mu)$ profile, Corollary~3's condition holds on only 7.62\% of cases, whereas the covariance/Cauchy--Schwarz bound in Corollary~4 holds on 12.94\% with zero violations. Crucially, across the 60 real converged points where Corollary~3 explained nothing, Corollary~4 holds without exception across all 30 toy, 15 digits at $T = 5$, and 15 digits at $T = 10$ seeds. All findings are fully reproducible using \texttt{verify\_proposition2\_corollary4.py} in the supplementary materials.

\section{Full Detail: Core Mechanism Validation Studies (i, ii, iii)}
\label{app:evidence}

\subsection{Shared Synthetic Landscape (Studies i--iii)}
Studies (i)--(iii) share one synthetic landscape, decoupled from any real architecture or data: $T=5$ windows, each with $K=50$ sub-components. One window is seeded as genuinely informative, with $c_{t,k}\sim\mathrm{Unif}(0.5,1.0)$; the other four get $c_{t,k}\sim\mathrm{Unif}(0,0.4)$; $a_{t,k}\sim\mathrm{Unif}(0.5,5.0)$ for every window and sub-component. $\beta_t(\gamma,\mu)$ uses $\beta_{\max}=1.0$, with $\gamma$ clamped to $[0.01,20]$ and $\mu$ to $[0,T{-}1]$. Both objectives are optimized via Adam (400 steps, finite-difference gradients, no autodiff dependency), from 3 $\mu$-initializations ($0.1$, $0.5$, $0.9$ of the window range---the middle one coinciding exactly with the true informative window) $\times$ 30 random landscape seeds, for 90 runs per objective per condition. \emph{Ratio}---mean converged objective at informative-vs.-elsewhere $\mu$-initialization---is the statistic reported below. Full per-seed distributions are independently reproducible from the standalone code in supplementary materials.

\textbf{(i) Full initialization-ablation detail.} Table~\ref{tab:functional}'s ``Rational (main)'' rows are study (i)'s own result, under the paper's $v_{t,k}$ form and sum-normalization. $H(p_A)$ alone converges to a similar objective value whether $\mu$ is initialized at the true window or elsewhere (ratio $0.93$). The loss carries essentially no signal distinguishing genuine informativeness from arbitrary placement, exactly as Proposition~1 predicts. The residual objective separates the two conditions sharply (ratio $27.11\times$). This objective-value \emph{ratio} across the informative-vs.-elsewhere initialization groups is what the paper's selection protocol (comparing losses across initializations, Section~\ref{sec:validation}) actually exploits.

\begin{figure}[h]
\centering
\includegraphics[width=0.8\linewidth]{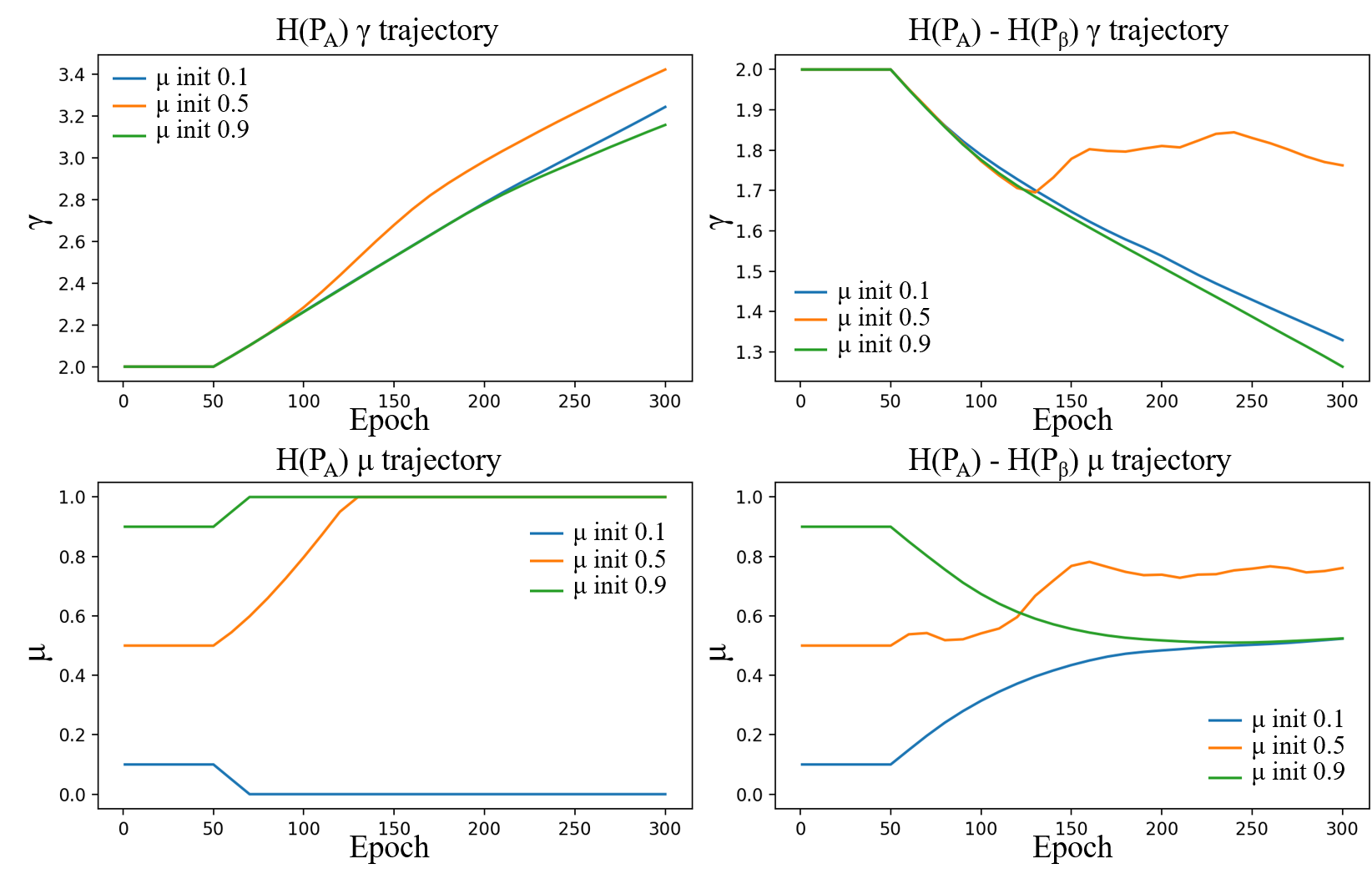}
\caption{Synthetic landscape: objective is blind to real signal under $H(p_A)$ alone, but not under the residual form.}
\label{fig:toy-landscape}
\end{figure}
\FloatBarrier

\textbf{(ii) Full functional-form and aggregation-scheme table.} Table~\ref{tab:functional} reports the rational, sigmoid, and logarithmic scoring forms and the softmax-gating variant under the same landscape and initialization protocol. Proposition~2's identity holds independently of these choices. The residual objective is negative for every reported form, and rational and logarithmic scores yield lower mean objectives from informative initialization, while sigmoid and softmax gating yield equal means across initialization groups. This equality does not by itself establish that the selected candidate is correct. The table therefore compares objective values and initialization dependence, rather than ranking selection accuracy across forms.

\begin{table}[h]
\caption{Full functional-form and aggregation-scheme generality table (study ii).}
\label{tab:functional}
\centering
\tablecaptionspace
\footnotesize
\setlength{\tabcolsep}{3pt}
\begin{tabular}{@{}llrrrr@{}}
\toprule
Family & Objective & obj(info.) & obj(else.) & ratio & $\gamma$(i/e) \\
\midrule
Rational (main) & $H(p_A)$ alone & +0.0047 & +0.0050 & 0.93 & 7.43/7.87 \\
Rational (main) & $H(p_A)-H(p_\beta)$ & $-0.3641$ & $-0.0134$ & 27.11 & 0.88/6.58 \\
Sigmoid & $H(p_A)$ alone & +1.1109 & +1.3517 & 0.82 & 6.52/6.37 \\
Sigmoid & $H(p_A)-H(p_\beta)$ & $-0.2502$ & $-0.2502$ & $1.00$ & 0.16/0.16 \\
Log & $H(p_A)$ alone & +0.0042 & +0.0046 & 0.91 & 7.33/7.74 \\
Log & $H(p_A)-H(p_\beta)$ & $-0.4139$ & $-0.0175$ & 23.67 & 0.86/7.05 \\
Softmax gating & $H(p_A)$ alone & +0.9940 & +1.5590 & 0.64 & 6.62/6.33 \\
Softmax gating & $H(p_A)-H(p_\beta)$ & $-0.3680$ & $-0.3680$ & $1.00$ & 0.14/0.14 \\
\bottomrule
\end{tabular}
\end{table}

\textbf{(iii) Full alternative regularizer table.} Table~\ref{tab:altreg} reports all four alternatives referenced by study (iii) in Table~\ref{tab:master}. Across all 90 runs (30 seeds across 3 initializations of $\mu$ spanning the window range, as defined in the shared landscape setup in Appendix~B), the final $\mu$ of the KL-divergence variant remains at its initial value. Rather than migrating toward the genuinely informative window, every run stays stuck at its initialization, matching the behavior observed under $H(p_A)$ alone. Similarly, $\gamma$ grows toward its unclamped ceiling (7.56 in informative regions and 7.91 elsewhere), mirroring the behavior under $H(p_A)$ alone (7.43 and 7.87, respectively). In contrast, the residual objective settles at moderate values of 0.88 in informative regions and 6.58 elsewhere. This failure stems from a structural issue: $D_{\mathrm{KL}}(p_A \parallel p_\beta)$ vanishes whenever $p_A = p_\beta$, which is precisely the arbitrary collapse point the regularizer ought to resist. Consequently, it offers no more protection than $H(p_A)$ alone.

\begin{table}[h]
\caption{Full alternative-regularizer table (study iii), including the KL-divergence variant. Same landscape and protocol as Table~\ref{tab:functional} (30 seeds $\times$ 3 $\mu$-inits).}
\label{tab:altreg}
\centering
\tablecaptionspace
\footnotesize
\setlength{\tabcolsep}{3pt}
\begin{tabular}{@{}lrrrr@{}}
\toprule
Objective & obj(info.) & obj(else.) & ratio & $\gamma$(i/e) \\
\midrule
$H(p_A)$ alone (reference) & +0.0047 & +0.0050 & 0.93 & 7.43/7.87 \\
$H(p_A)-H(p_\beta)$ (ours, reference) & $-0.3641$ & $-0.0134$ & 27.11 & 0.88/6.58 \\
$H(p_A) + 1.0\,(\gamma/\gamma_{\max})^2$ & +0.1004 & +0.1137 & 0.88 & 5.27/5.65 \\
$H(p_A)$, tighter $\gamma$ clamp ($\gamma_{\max}{=}2.0$ vs.\ $20.0$) & +0.3830 & +0.4543 & 0.84 & 2.00/2.00$^\ddagger$ \\
$H(p_A) + 5.0\sum_t(p_{\beta,t}-1/T)^2$ & +1.3691 & +2.1906 & 0.62 & 0.15/5.51 \\
$H(p_A) - \mathrm{KL}(p_A\Vert p_\beta)$ & +0.0040 & +0.0048 & 0.83 & 7.56/7.91 \\
\bottomrule
\end{tabular}
\end{table}

$^\ddagger$``gamma\_clip'' uses a tighter hard ceiling ($\gamma_{\max}{=}2.0$ instead of $20.0$) by construction, so both columns sit at the clamp (frac.\ at clamp $=1.00$) rather than reflecting free convergence. The ratios summarize objective differences between initialization groups; they are not a common performance scale across these different objectives.

\subsection{Reproducibility statement}

Studies (i)--(iii) are all reproduced by \texttt{toy\_entropy\_bilevel.py} in supplementary materials, standalone-runnable with only numpy: running it with no flags reproduces study (i) and the alternative-regularizer table (study (iii), Table~\ref{tab:altreg}), while the \texttt{-{}-families} flag reproduces the functional-form/aggregation table (study (ii), Table~\ref{tab:functional}). Each completes in well under a minute.

\section{Full Detail: Class-Imbalance Identification and Downstream Reweighting (Study iv)}
\label{app:digits}
\label{app:confidence}

\subsection{Experimental pipeline}

Study (iv) runs two separate models per seed, linked only by the output of Phase~1 (Algorithm~2). Phase~1 pre-computes fixed signals once (Stage~A), optimizes them from $T$ independent $\mu$-initializations (Stage~B), and combines the results by majority vote (Stage~C). Phase~2 never sees $c_{t,k}$, $a_{t,k}$, $\gamma$, $\mu$, or $\beta_t$ and never updates Phase~1. It receives only the identified class and confidence score through a \texttt{sample\_weight} vector.

\begin{algboxhere}{2}{Exact computational order for one seed in study (iv) (elaborating Algorithm~1). \label{alg:phase4detail}}
\footnotesize
\textbf{Phase 1 --- mixture identification (Stages A--C; never touches Phase 2's coefficients)}
\vspace{1pt}\hrule\vspace{2pt}
\textit{Stage A -- fixed batch pre-computation (runs once; not a loop)}

\algln $(X_p^{tr},X_p^{val},y_p^{tr},y_p^{val}) \gets$ 60/40 split of $(X_{\text{train}},y_{\text{train}})$

\algln \texttt{probe.fit}$(X_p^{tr},y_p^{tr})$ \hfill\textit{// one-vs-rest logistic regression, discarded after this stage}

\algln $\text{proba} \gets \texttt{probe.predict\_proba}(X_{\text{train}})$ \hfill\textit{// batch call on all $N$ rows at once}

\algln $\text{uncertainty}_k \gets 1-\max_c \text{proba}_{k,c}$ for every training row $k$

\algln $\text{class\_count}(t) \gets |\{k: y_k=t\}|$; \ $\text{scarcity}(t) \gets \max_{t'}\text{class\_count}(t')\,/\,\text{class\_count}(t)$

\algln \textbf{for} $t=1,\dots,T$: \ $X_t \gets$ rows of $X_{\text{train}}$ with label $t$; fit \texttt{NearestNeighbors} on $X_t$; $a_{t,k} \gets$ mean of its 4 nearest same-class distances $+10^{-3}$

\algln $c_{t,k} \gets \text{scarcity}(t)\cdot(0.5+\text{uncertainty}_k)$ for every $k$ with true class $t$

\algln \textbf{return} frozen vectors $\{c_k\},\{a_k\}$, one pair per training row -- reused unchanged in every Stage-B call below

\vspace{2pt}\hrule\vspace{2pt}
\textit{Stage B -- optimization (Algorithm~1; run independently for $i=1,\dots,T$, once per initialization $\mu_0^{(i)}$)}

\algln $\theta \gets (\gamma_0,\mu_0^{(i)})$

\algln \textbf{while} not converged (\hspace{-0.2em}$\leq400$ Adam steps) \textbf{do}

\hspace{1.5em}\algln $\beta_t \gets e^{-\gamma(t-\mu)^2}$ for all $t$

\hspace{1.5em}\algln $v_t \gets \sum_{k:\,t_k=t} c_k\,\beta_t/(\beta_t+a_k)$ for all $t$ \hfill\textit{// consumes frozen $c_k,a_k$}

\hspace{1.5em}\algln $p_{A,t}\gets v_t/\sum_{t'}v_{t'}$, \ $p_{\beta,t}\gets\beta_t/\sum_{t'}\beta_{t'}$; \ $D\gets H(p_A)-H(p_\beta)$

\hspace{1.5em}\algln $\theta \gets \theta-\eta\nabla_\theta L$ (finite-difference gradient), $L=L_1$ or $L_2$

\algln \textbf{end while}

\algln $r_t \gets v_t/\beta_t$ for all $t$; \ $t^{*(i)}\gets\arg\max_t\beta_t$; \ $t'\gets\arg\max_{t\neq t^{*(i)}} r_t$

\algln \textbf{return} $t^{*(i)}$, \ $m^{(i)}\gets(r_{t^{*(i)}}-r_{t'})/(r_{t^{*(i)}}+r_{t'}+\epsilon)$

\vspace{2pt}\hrule\vspace{2pt}
\textit{Stage C -- combine the $T$ independent restarts (majority vote; not an optimization)}

\algln identified class $\gets$ most common value among $t^{*(1)},\dots,t^{*(T)}$

\algln $\bar m \gets$ mean of $m^{(i)}$ over only the $i$ whose $t^{*(i)}=$ identified class

\algln confidence $\gets\mathrm{clip}(\bar m/s,\,0,\,1)$, \ $s=0.25$

\vspace{3pt}\hrule\vspace{3pt}
\textbf{Phase 2 --- downstream classifier (separate model; repeated once per condition $\in\{$uniform, fixed$\times5$, conf-scaled, oracle$\}$)}
\vspace{1pt}\hrule\vspace{2pt}

\algln $w \gets \mathbf{1}_N$; \ factor $\gets 1$ (uniform), $5$ (fixed$\times5$), $1{+}4\cdot\text{confidence}$ (conf-scaled), or $5$ on the true minority class (oracle)

\algln $w_k \gets$ factor for every row $k$ whose true label is the identified class (oracle: the true minority class); all other $w_k$ stay $1$

\algln \texttt{clf.fit}$(X_{\text{train}},y_{\text{train}},\text{sample\_weight}{=}w)$ \hfill\textit{// one batch call; separate object from \texttt{probe}}

\algln $\hat y \gets \arg\max_c \texttt{clf.predict\_proba}(X_{\text{test}})$ (batch, all held-out rows at once)

\algln \textbf{return} $\text{recall\_score}(y_{\text{test}},\hat y,\text{labels}{=}[\text{minority\_idx}],\text{average}{=}\text{"macro"})$ \hfill\textit{// minority recall (Table~\ref{tab:digitsB})}

\algln \textbf{also return} $\text{accuracy\_score}(y_{\text{test}},\hat y)$ \hfill\textit{// overall accuracy, all classes (Table~\ref{tab:digitsB})}
\end{algboxhere}

We use classes 0--4 from scikit-learn's digits dataset ($T{=}5$), whose examples are real $8\times8$ grayscale images. We retain only 8\% of digit~3's training examples (about 10 images) and leave the other classes intact. The task is to identify the undisclosed minority class and determine whether acting on that selection through oversampling improves its recognition by a separately trained classifier.

Each training example receives a class-level scarcity signal and an example-level uncertainty signal from a probe classifier fitted on the internal probe-training split:
\[
\text{scarcity}(t) := \frac{\max_{t'}\text{count}(t')}{\text{count}(t)},
\qquad
\text{uncertainty}_k := 1 - \max_c P(c\mid x_k).
\]
Their product defines informativeness,
\[
c_{t,k} = \text{scarcity}(t)\cdot(0.5+\text{uncertainty}_k).
\]
Here $a_{t,k}$ is the mean distance to example $k$'s four nearest same-class neighbors. The shared computational pipeline in Section~\ref{sec:validation} aggregates
\[
v_{t,k}(\beta_t) = \frac{c_{t,k}\,\beta_t}{\beta_t + a_{t,k}}, \qquad v_t = \sum_k v_{t,k}(\beta_t).
\]
We repeat the experiment over 80 random splits.

\FloatBarrier

\subsection{Identification and downstream value at 8\% scarcity}

Table~\ref{tab:digitsB} separates identification from its downstream value. Phase~1 reports the fraction of seeds in which the majority vote across five initializations (Algorithm~2) identifies the true minority class, normalized by the $H(p_A)$-alone rate. Phase~2 retrains a separate classifier after up-weighting the selected class and reports held-out minority recall and overall accuracy. The oracle instead up-weights the known minority class and serves only as an upper-bound reference.

\begin{table}[h]
\caption{Digits ($T{=}5$; minority class retained at 8\%, approximately 10 training images; 80 seeds and five $\mu$-initializations per objective). Phase~1 reports the true-minority selection rate relative to $H(p_A)$ alone. Phase~2 reports minority recall and overall accuracy after separately retraining with a fixed $5\times$ weight on the class selected by each objective. The 2\% variant appears in Appendix~\ref{app:digits-hard}.}
\label{tab:digitsB}
\centering
\tablecaptionspace
\small
\begin{tabular}{@{}lrrr@{}}
\toprule
Phase 1 (outer loop) & $H(p_A)$ alone & Residual (ours) & \\
\midrule
Relative selection improvement (true minority class) & $1.00\times$ (0.775) & $1.21\times$ (0.938) & \\
\multicolumn{4}{l}{Two-proportion $z$-test on raw selection rate: $z=2.929$, $p=0.003$ (significant)} \\
\midrule
Phase 2 (downstream) & Minority recall & Accuracy & \\
\midrule
Uniform (no emphasis) & 0.8466 & 0.9635 & \\
Residual-selected class, fixed $5\times$ & 0.8727 & 0.9687 & \\
$H(p_A)$-selected class, fixed $5\times$ & 0.8670 & 0.9675 & \\
Oracle (true minority class emphasized) & 0.8743 & 0.9690 & \\
\multicolumn{4}{l}{Paired $t$-test, residual-identified vs.\ uniform (minority recall): mean diff $=+0.0261$, $t=9.959$, $p<0.0001$} \\
\bottomrule
\end{tabular}
\end{table}

The residual objective improves selection from 0.775 to 0.938 ($1.21\times$, $p{=}0.003$) and removes the initialization-locked behavior predicted by Proposition~1. Under $H(p_A)$ alone, four of five initializations remain at their starting classes in every seed. Only $\mu{=}2$ moves, reaching the minority class in 62/80 seeds. Under the residual objective, $\mu{=}4$ never remains at its starting class (0/80 versus 80/80) and reaches the minority class in every seed.  $\mu{=}2$ also always moves, with its five errors arising from data-dependent votes for class~1 rather than boundary retention. Correspondingly, $D\leq0$ at 396/400 converged residual runs (mean $D=-0.20$). The four positive values are at numerical precision ($\leq1.4\times10^{-4}$). Acting on the residual selection with fixed $5\times$ weighting raises minority recall from 0.8466 to 0.8727 and approaches the oracle value of 0.8743 ($p<0.0001$ versus uniform), showing that the improved identification transfers to a separately trained classifier.

The margin adds a second, distinct payoff. It controls how strongly the selected class is acted upon. We set $s{=}0.25$ by inspecting the 8\%-scarcity margins and use confidence $=\mathrm{clip}(m/s,0,1)$. The margins separate without overlap: $[0.26,0.49]$ for the 75 correctly identified seeds and $[-0.42,-0.32]$ for the five misidentified seeds. Thus, confidence saturates at 1 for every correct selection and clips to 0 for every error. Even when the outer loop selects the wrong class, $r_t$ still favors the true minority class over that selection, making $m$ negative and flagging the decision as unreliable.

Confidence-scaled reweighting consequently obtains mean minority recall 0.8732, compared with 0.8727 for fixed $5\times$ weighting ($p{=}0.159$) and 0.8466 for uniform training ($p<0.0001$; Table~\ref{tab:confreweight}). Its improvement over the fixed factor is not significant because only five seeds differ, but it removes both cases in which fixed weighting falls below uniform (0/80 harmful cases versus 2/80). At this scarcity level, improved identification supplies the primary downstream gain. The margin provides downside protection when identification fails.

\subsection{Hard-task variant (Table~\ref{tab:master}, row iv)}
\label{app:digits-hard}
\label{app:confidence-hard}

We repeat Phase~1 at 2\% scarcity using the same 15 seeds, splits, and code path. Selection rates tie at $8/15$ for both objectives ($1.00\times$), unlike the 8\% setting, where the residual objective achieves a significant $1.21\times$ improvement ($p{=}0.003$).

Table~\ref{tab:confreweight} shows the downstream consequence. Fixed $5\times$ weighting harms 5/15 seeds, whereas confidence scaling harms none and raises mean minority recall from 0.6909 to 0.7018. This yields a significant advantage over both fixed weighting ($p{=}0.023$) and uniform training (recall 0.6776, $p{=}0.012$). Confidence scaling cannot correct the selected class. On a misidentified seed, confidence clips to 0 and the factor becomes exactly 1, reproducing uniform training. Its role is to prevent a low-margin error from being amplified downstream.

\subsection{Scope and reproducibility}

The scale $s{=}0.25$ was selected by inspection rather than an automated calibration procedure and was not re-tuned at 2\% scarcity. It is a fixed choice, not an optimized or independently validated hyperparameter. The experiments use public packages and scikit-learn's digits dataset. \path{second_application_digits_imbalance.py B_T5_sharp} reproduces identification (defaults: \path{minority_frac=0.08}, \path{n_seeds=80}; hard variant: 0.02 and 15). \path{sixth_application_confidence_reweighting.py} reproduces downstream reweighting and the scale sweep with the same settings.

In summary, residual entropy can improve identification over Shannon entropy, and downstream value depends on alignment with informativeness. Well-aligned selections benefit a separately trained classifier, while the derived margin prevents poorly aligned selections from being amplified.

\section{Full Detail: Coupled, Non-Stationary MoE Experiments (Studies v--vi)}
\label{app:moe}

\subsection{Shared setup and computational procedure}

Both studies use $T{=}6$ independently trained softmax-classifier experts ($W_e\in\R^{64\times 10}$, bias $b_e$) on scikit-learn's digits dataset (1797 real $8{\times}8$ images, 10 classes; 60/20/20\% train/validation/test per seed). The $K{=}10$ sub-components are the classes: $c_{e,k}$ is expert $e$'s mean predicted probability of the true class among validation examples of class $k$, recomputed after every step, and $a_{e,k}{=}0.3$ is shared across experts and classes. We interpret $\beta_e(\gamma,\mu)$ as population-level routing emphasis and set expert $e$'s learning-rate multiplier to $\mathrm{clip}(T\beta_e/\sum_{e'}\beta_{e'},\,\mathrm{floor},\,5.0)$. The favored expert therefore trains faster and tends to achieve higher measured accuracy at the next step, creating a rich-get-richer loop. Expert~3 ($e{=}2$ zero-indexed) receives a measured advantage through 25 warm-start gradient steps. The other five begin from small random initializations. The outer loop uses Adam with finite-difference gradients for 150 steps. A held-out test split, unused in training or in computing $c_{e,k}$, evaluates the final gate-weighted ensemble.

Unlike study (iv) (Algorithm~2), $c_{e,k}$ is not frozen before optimization. The experts and $(\gamma,\mu)$ are updated jointly, with each step feeding the next. Selection across $\mu$-initializations also differs from study (iv)'s majority vote. Because no independent ground-truth signal is available at run time, we retain the run with the lowest final objective. Algorithm~3 gives the full-batch version used in study (v). Study (vi) changes only how training data are exposed to the experts.

\begin{algboxhere}{3}{Study (v)'s exact computational order, one seed, one $\mu$-initialization (contrast with Algorithm~2: here Stage A is one non-stationary joint loop, not a frozen pre-computation). \label{alg:moedetail}}
\footnotesize
\textbf{Stage A --- joint inner (expert) + outer $(\gamma,\mu)$ loop (the only loop; $c_{e,k}$ evolves every step)}
\vspace{1pt}\hrule\vspace{2pt}

\algln $(X_{tr},y_{tr},X_{val},y_{val},X_{te},y_{te}) \gets$ 60/20/20 split; init $W_e{\sim}\mathcal N(0,0.01^2)$, $b_e{\gets}0$ for $e=1,\dots,T$

\algln \textbf{if} $e=E_{\text{TRUE}}$: run 25 full-batch gradient steps on $(X_{tr},y_{tr})$ \hfill\textit{// warm-start advantage, once}

\algln $\theta \gets (\gamma_0{=}2.0,\ \mu_0)$; \ $c_{e,k} \gets$ \texttt{batched\_c\_ek}$(W,b)$ \hfill\textit{// initial snapshot, batched over all $T$ experts at once}

\algln \textbf{for} step $=1,\dots,150$ \textbf{do}

\hspace{1.5em}\algln $\beta_e \gets \beta_{\max}e^{-\gamma(e-\mu)^2}$ for all $e$; \ lr\_mult$_e \gets \mathrm{clip}(T\beta_e/\textstyle\sum_{e'}\beta_{e'},\,0.05,\,5.0)$

\hspace{1.5em}\algln $(W,b) \gets$ one gradient-descent step on \emph{every} expert's softmax classifier on $(X_{tr},y_{tr})$, expert $e$'s own learning rate $=0.5\cdot\text{lr\_mult}_e$ \hfill\textit{// inner: real neural-net-style training}

\hspace{1.5em}\algln $c_{e,k} \gets$ \texttt{batched\_c\_ek}$(W,b)$ on $(X_{val},y_{val})$ \hfill\textit{// re-measured from the just-updated weights, every step}

\hspace{1.5em}\algln $v_e \gets \sum_k c_{e,k}\beta_e/(\beta_e+a_{\text{shared}})$; \ $p_{A,e}\gets v_e/\sum v$, $p_{\beta,e}\gets\beta_e/\sum\beta$

\hspace{1.5em}\algln $L \gets H(p_A)$ (mode: alone), or $H(p_A){-}H(p_\beta)$ (mode: residual), or $H(p_A){+}\alpha T\sum_e p_{\beta,e}^2$ (mode: switch-aux)

\hspace{1.5em}\algln $\theta \gets \theta - \eta\,\text{Adam}(\nabla_\theta L)$, finite-difference gradient \emph{holding this step's $c_{e,k}$ fixed} (only $\beta_e(\theta)$ is differentiated); clip $\gamma\in[0.01,20]$, $\mu\in[0,T{-}1]$

\algln \textbf{return} $\theta^*$, final $(W,b)$, final objective $L^*$, and win $\gets\mathrm{round}(\mathrm{clip}(\mu^*,0,T{-}1))$

\vspace{2pt}\hrule\vspace{2pt}
\textbf{Stage B --- select across the 3 independent $\mu$-initializations (lowest objective, not majority vote)}
\vspace{1pt}\hrule\vspace{2pt}

\algln run Stage A independently for $\mu_0\in\{0.1(T{-}1),\,0.5(T{-}1),\,0.9(T{-}1)\}$ $\to$ 3 candidates, each with its own $(\theta^*,W,b,L^*,\text{win})$

\algln selected $\gets \arg\min$ over the 3 candidates by $L^*$ \hfill\textit{// no ground truth available at run time, unlike study (iv)}

\algln selection correct $\gets \mathbb{1}[\text{selected.win}=E_{\text{TRUE}}]$ \hfill\textit{// only used for reporting; not fed back into the run}

\vspace{2pt}\hrule\vspace{2pt}
\textbf{Downstream --- gate-weighted ensemble on the held-out test split (no separate retraining)}
\vspace{1pt}\hrule\vspace{2pt}

\algln gate$_e \gets \beta_e(\theta^*)/\sum_{e'}\beta_{e'}(\theta^*)$ from the selected candidate's final $\theta^*$ \hfill\textit{// same $\beta$ that drove training, reused as-is}

\algln $P_e(x) \gets \mathrm{softmax}(x W_e + b_e)$ for every expert $e$, on $X_{te}$ (the selected candidate's final, already-trained $W,b$ -- \emph{no new model is fit here})

\algln $P_{\text{ens}}(x) \gets \sum_e \text{gate}_e\,P_e(x)$; \ $\hat y \gets \arg\max_c P_{\text{ens}}(x)$

\algln \textbf{return} test accuracy $=\mathrm{mean}(\hat y = y_{te})$ (Table~\ref{tab:moe-floor-sweep})
\end{algboxhere}

\subsection{Residual-entropy diagnostic in the full-batch regime (Study v)}
\label{app:moe-fair}

Algorithm~3 gives the base three-initialization procedure. The fair protocol below instead runs Stage~A from all six candidate-centered initializations and retains the lowest-objective run.

\begin{table}[h]
\caption{Study (v), floor sweep under the fair protocol (one $\mu$-init per candidate window, best-of-6 by lowest final objective, 9 seeds per floor, $E_{\text{TRUE}} = 2$, noisy hard-task variant). An expert is lower-bounded when its learning-rate multiplier $\mathrm{lr\_mult}_e = \mathrm{clip}(\beta_e/\bar{\beta},\,\mathrm{floor},\,5.0)$ sits at the lower bound, i.e.\ $\mathrm{lr\_mult}_e = \mathrm{floor}$. Only floor$=0$ is literal starvation (exact zero gradient). For floor$>0$ these experts still train, rate-capped at floor.}
\label{tab:moe-floor-sweep}
\centering
\tablecaptionspace
\small
\begin{tabular}{@{}lrrrrl@{}}
\toprule
 & \multicolumn{2}{c}{$H(p_A)$ alone} & \multicolumn{2}{c}{Residual (ours)} & \\
floor & lower bdd.\ (of 6) & mean $\gamma_f$ & lower bdd.\ (of 6) & mean $\gamma_f$ & clip effect \\
\midrule
$0.00$ & $4.00$ & $6.76$ & $0.00$ & $0.01$ & literal freeze \\
$0.05$--$0.07$ & $5.00$ & $6.77$ & $0.00$ & $0.01$ & trickle \\
$0.10$--$0.30$ & $5.00$ & $6.79$ & $4.00$ & $\approx4.06$ & trickle \\
$1.00$ & $5.00$ & $6.82$ & $4.00$ & $4.08$ & trickle \\
\bottomrule
\end{tabular}
\end{table}

Table~\ref{tab:moe-floor-sweep} asks whether Corollary~2's uniform, non-starving regime is reachable in a trained MoE. With $H(p_A)$ alone, the answer does not depend on floor: 5/6 experts remain lower-bounded for every floor$>0$, and 4/6 are literally starved at floor$=0$. Residual entropy instead produces a sharp transition. At floor$\leq0.07$, $\gamma\approx0$ and no expert is lower-bounded. At floor$\geq0.10$, $\gamma\approx4.1$ and 4/6 remain lower-bounded. Thus the residual term turns Corollary~2 into an observable operating boundary that Shannon entropy alone does not provide.

Because the boundary shifts with task noise (Table~\ref{tab:floor-threshold-noise}), we next test whether changing floor during training can outperform choosing one side from the start. At each of seven noise levels, we compare static floor values of $0.05$ and $1.0$ with schedules that switch between them at step 75 ($n{=}25$ per cell).

\label{app:selection-metric-caveat}

Additional schedules give the same result: on the noisy task, starting low yields residual accuracy near $0.804$, whereas starting high or dipping low only mid-training yields $0.783$--$0.784$. $H(p_A)$ alone remains at $0.775$--$0.777$ with 5/6 experts lower-bounded throughout. The initial side of the residual threshold, not the final floor value, determines the learned regime.

\begin{table}[!htbp]
\caption{Study (v), floor-adjustment strategy comparison across seven noise levels (residual objective, fair protocol, $n{=}25$ per cell). floor$_{\text{low}}=0.05$, floor$_{\text{high}}=1.0$. Both schedules switch at step $75$ of $150$.}
\label{tab:moe-floor-strategy}
\centering
\tablecaptionspace
\footnotesize
\setlength{\tabcolsep}{4pt}
\begin{tabular}{@{}lrrrrl@{}}
\toprule
Noise & static\_low & static\_high & low$\to$high@75 & high$\to$low@75 & Winner \\
\midrule
$0.0$ & $0.957$ & $0.963$ & $0.959$ & $0.963$ & high ($+0.5$pt) \\
$0.5$ & $0.922$ & $0.916$ & $0.922$ & $0.916$ & low ($+0.5$pt) \\
$1.0$ & $0.804$ & $0.783$ & $0.804$ & $0.783$ & low ($+2.1$pt) \\
$1.5$ & $0.644$ & $0.643$ & $0.644$ & $0.643$ & low ($+0.1$pt) \\
$2.0$ & $0.537$ & $0.537$ & $0.537$ & $0.537$ & tie \\
$2.5$ & $0.442$ & $0.441$ & $0.442$ & $0.441$ & low ($+0.1$pt) \\
$3.0$ & $0.373$ & $0.357$ & $0.373$ & $0.357$ & low ($+1.6$pt) \\
\bottomrule
\end{tabular}
\end{table}

Table~\ref{tab:moe-floor-strategy} shows that each schedule follows its starting floor to three decimal places: low$\to$high matches static\_low, and high$\to$low matches static\_high. The regime is therefore selected early and is not repaired by a later switch. The practical use of the threshold is to diagnose the deployment data and start training on the accuracy-preferred side, rather than to adjust floor after commitment. Scripts: \path{floor_threshold_noise_sweep.py} and \path{floor_strategy_search.py}.

\subsection{Comparison with a standard load-balancing objective}
\label{app:moe-auxloss}

This comparison asks whether standard load balancing provides the same information as residual entropy. Switch Transformer's auxiliary loss (Fedus et al., 2022, Eq.~4) is $L_{\text{aux}} = \alpha N \sum_i f_i P_i$ ($f_i$: hard-dispatch fraction; $P_i$: mean routing probability). In our population-level gate, $f_i=P_i$, giving $L_{\text{switch}} = H(p_A) + \alpha T \sum_e p_{\beta,e}^2$.

\begin{table}[h]
\caption{Study (v) extended with a Switch-Transformer-style load-balancing loss (App.~\ref{app:moe-auxloss}), 6-candidate fair protocol (25 seeds, best-of-6 by lowest final objective), same protocol as Table~\ref{tab:moe-floor-sweep}.}
\label{tab:switchaux}
\centering
\tablecaptionspace
\small
\begin{tabular}{@{}lrr@{}}
\toprule
 & Lower bdd.\ (of 6) & Ensemble test acc. \\
\midrule
$H(p_A)$ alone & 5.00 & $0.9628\pm0.0092$ \\
Switch-style aux ($\alpha{=}0.01$) & 5.00 & $0.9631\pm0.0094$ \\
Residual (ours) & 0.00 & $0.9570\pm0.0103$ \\
\bottomrule
\end{tabular}
\end{table}
\medskip
At the default $\alpha=0.01$, the Switch-style objective behaves like $H(p_A)$ alone. Both leave 5/6 experts lower-bounded and attain indistinguishable ceiling-level accuracy ($0.9631$ vs.\ $0.9628$; Table~\ref{tab:switchaux}). Residual entropy removes lower-bounding entirely, although its $0.9570$ accuracy is not higher in this clean ceiling regime. It is also the only objective to select the genuinely advantaged expert at all (2/25 vs.\ 0/25), a small count that supports the arbitrary-placement distinction but is not an accuracy claim. The comparison therefore separates the intended roles of the objectives. Clean-task accuracy masks routing differences, while residual entropy changes both the starvation regime and, occasionally, which expert is identified.

\begin{table}[h]
\caption{Load-balancing coefficient sweep, 6-candidate fair protocol (10 seeds each, best-of-6 by lowest final objective), floor$=0.05$ (never $0$, so counts below are lower-bounded, not literal starvation). Lower-bounding is fixed monotonically by a large enough $\alpha$.}
\label{tab:alphasweep}
\centering
\tablecaptionspace
\small
\begin{tabular}{@{}lr@{}}
\toprule
$\alpha$ & Lower bdd.\ (of 6) \\
\midrule
0.01 & 5.00 \\
0.1  & 5.00 \\
1.0  & 0.00 \\
10.0 & 0.00 \\
\bottomrule
\end{tabular}
\end{table}

Increasing $\alpha$ confirms that the Switch-style loss performs its intended balancing function: lower-bounding falls from 5/6 to 0/6 once $\alpha\geq1.0$ (Table~\ref{tab:alphasweep}). The point is therefore not that load balancing fails, but that it answers a different question. Its response is a monotonic strength threshold on gate concentration. Residual entropy supplies a data-dependent boundary tied to the safety analysis and is designed to distinguish informative from arbitrary concentration.

\subsection{Streaming-data robustness check (Study vi)}
\label{app:bandit}

This study reuses study (v)'s trained-system design (Appendix~\ref{app:moe}: $T{=}6$ real softmax-classifier candidates, $K{=}10$ classes, $a_{e,k}{=}0.3$ shared, population-level gate) with one structural change. Instead of seeing the full real training set every outer step, the digits training pool is shuffled once per seed and split into 15 sequential chunks. Outer step $r$ exposes only chunk $(r\bmod 15)$, so each candidate's weights update incrementally on a different data slice each round rather than via a repeated full-batch gradient step. Candidate 3 ($E_{\text{TRUE}}{=}2$ zero-indexed) receives the same measured warm-start advantage at the same interior position. The base implementation uses 150 outer rounds, 3 $\mu$-initializations, and 25 seeds. The comparisons below use the candidate-centered best-of-6 fair protocol stated in their captions.

\label{app:bandit-fair}

\begin{table}[h]
\caption{Study (vi), floor sweep under the fair protocol (one $\mu$-init per candidate window, best-of-6 by lowest final objective, 9 seeds per floor, $E_{\text{TRUE}}{=}2$, noisy hard-task variant): lower-bounded-candidate count and mean converged $\gamma_f$, both objectives. Same lower-bounded/literal-starvation distinction as Table~\ref{tab:moe-floor-sweep} (floor$=0$ only).}
\label{tab:bandit-floor-sweep}
\centering
\tablecaptionspace
\small
\begin{tabular}{@{}rrrrrl@{}}
\toprule
 & \multicolumn{2}{c}{$H(p_A)$ alone} & \multicolumn{2}{c}{Residual (ours)} & \\
floor & lower bdd.\ (of 6) & mean $\gamma_f$ & lower bdd.\ (of 6) & mean $\gamma_f$ & clip effect \\
\midrule
$0.000$ & $4.00$ & $6.76$ & $0.00$ & $0.01$ & literal freeze \\
$0.062$ & $5.00$ & $6.77$ & $0.44$ & $0.46$ & trickle \\
$0.070$ & $5.00$ & $6.77$ & $3.56$ & $3.61$ & trickle \\
$0.100$--$1.000$ & $5.00$ & $6.78$--$6.82$ & $4.00$ & $\approx4.07$--$4.08$ & trickle \\
\bottomrule
\end{tabular}
\end{table}

The streaming experiment asks whether the residual threshold survives when the information signal changes with each data chunk. It does: $H(p_A)$ alone again leaves 5/6 candidates lower-bounded for every floor$>0$, whereas residual entropy transitions sharply between floor $0.062$ (0.44 lower-bounded on average) and $0.070$ (3.56; Table~\ref{tab:bandit-floor-sweep}). This reproduces the full-batch transition under a different data-exposure process, showing that the boundary is not an artifact of repeatedly training on the same batch. Accuracy follows the resulting regime rather than the numerical floor value. Floor $0.07$ and $1.0$ are indistinguishable ($p{=}0.347$), while floor$=0$ significantly exceeds both ($p{=}0.0001$ each). Script: \texttt{collapse\_phase\_transition.py} (\texttt{part\_I5}).

\begin{table}[h]
\caption{Study (vi), fair protocol (one init per window, best-of-6 by lowest final objective, 9 seeds). Same three floor/noise conditions as study (v)'s corresponding ablation (Appendix~\ref{app:moe-fair}).}
\label{tab:bandit-fair}
\centering
\tablecaptionspace
\small
\begin{tabular}{@{}lrr@{}}
\toprule
 & $H(p_A)$ alone & Residual (ours) \\
\midrule
Default (floor$=0.05$, clean) & $0.960\pm0.008$ & $0.960\pm0.010$ \\
Hard variant (floor$=0$, noisy) & $0.756\pm0.016$ & $0.805\pm0.017$ \\
Floor$=1$ (clean) & $0.960\pm0.008$ & $0.962\pm0.007$ \\
\bottomrule
\end{tabular}
\end{table}

On clean data, both objectives remain at the accuracy ceiling regardless of their different starvation behavior (Table~\ref{tab:bandit-fair}). Under the noisy hard condition, however, residual entropy raises accuracy from $0.756$ to $0.805$. The streaming study therefore establishes two points. The residual safety transition persists under non-stationary data, and preventing collapse becomes operationally valuable when task difficulty removes the clean-data ceiling. Scripts: \texttt{bandit\_hardtask\_test.py}, \texttt{hardtask\_floor\_ablation.py}, and \texttt{bandit\_floor\_starved\_sweep.py}.

\subsection{Separating starvation prevention from correct selection}

We finally ask which property drives ensemble accuracy here: keeping all experts trainable or identifying the advantaged expert. When starvation status is held fixed, choosing the correct expert provides no measurable advantage. Accuracy is unchanged whether the gate is correct or incorrect when starvation is prevented ($0.808$ vs.\ $0.806$, $p{=}0.628$; $n{=}54$) and when starvation remains ($0.769$ vs.\ $0.776$, $p{=}0.351$; $n{=}27$). The practical comparison is even clearer. A model that prevents starvation but selects the wrong expert reaches $0.807$, whereas a model that selects the right expert but allows starvation reaches $0.769$. In this ensemble setting, preserving expert participation matters more than gate correctness.

This result does not imply that correct selection has no value. Here, the experts are interchangeable and their outputs are averaged, so an ordering error is absorbed by the ensemble. Appendix~\ref{app:fungibility} changes those two conditions and shows that correct ordering becomes beneficial when candidates differ and the selected candidate is acted upon directly. Script: \texttt{collapse\_phase\_transition.py} (\texttt{part\_I6}).

\subsection{Scope and reproducibility}

The gate $\beta_e(\gamma,\mu)$ is a population-level routing emphasis, not a per-token top-$k$ gate in a production sparse MoE \citep{shazeer2017,lepikhin2021,fedus2022}. The experiments deliberately isolate the same rich-get-richer feedback loop in a simpler system. Study (vi) tests transfer to sequential data exposure, not regret-minimizing bandit performance. It uses no exploration bonus or regret bound and makes no comparison claim against UCB, Thompson sampling, or noisy top-$k$ gating. We likewise do not propose residual entropy as a replacement for production load-balancing losses. The comparison shows that sufficiently strong load balancing can eliminate lower-bounding, whereas residual entropy additionally exposes a data-dependent transition and addresses arbitrary-placement selection.

The full-batch implementation, including the vectorized expert loop and Switch-style comparison, is \path{fourth_application_moe_routing.py}. Its \path{run_alpha_sensitivity()} function and \path{--alpha-sensitivity} flag reproduce the coefficient sweep. \path{floor_threshold_noise_sweep.py} and \path{floor_strategy_search.py} reproduce the noise-dependent threshold and floor schedules. The streaming implementation is \path{fifth_application_bandit_sequential_selection.py}. All use public packages and scikit-learn's public digits dataset and are standalone-runnable.

In summary, across full-batch and streaming MoE training, residual entropy reveals a data-dependent safety threshold that Shannon entropy does not, making floor a meaningful control for lower-bounding and the accuracy-preferred regime, while correct selection alone remains ineffective when interchangeable experts are evaluated through an ensemble.

\section{Full Detail: Downstream Value of Correct Ordering}
\label{app:fungibility}

\subsection{Experimental question and design}

Corollary~2 guarantees safety, whereas Corollaries~3--4 guarantee that trust and informativeness order candidates consistently. Neither guarantee implies the other (Section~\ref{sec:fix}). Appendix~\ref{app:bandit} found that correct gate selection did not improve ensemble accuracy even though starvation prevention did. We test two explanations for that null. First, a gate-weighted ensemble may absorb a ranking error because several experts still contribute to the final prediction. Second, experts trained on the same data may become similar enough that choosing the correct one has little effect on the output. Expert redundancy is not unique to our setting. Prior work has found substantial redundancy in trained sparse MoEs \citep{li2024}, while DeepSeekMoE explicitly targets shared knowledge across routed experts to encourage greater specialization \citep{dai2024}. The experiments below test whether this interchangeability explains our null. The first control consumes the selection directly, the second applies the same intervention to the original MoE, and the final experiment manipulates interchangeability itself. 

\subsection{Directly consuming the selection}

Under a gate-weighted average, mis-ranking causes no visible harm because the ensemble simply absorbs the error. However, when the pipeline actually consumes the selection, by up-weighting the chosen candidate class and retraining (study (iv)'s digits pipeline, paired by seed, $n{=}50$, $T{=}5$, factor $5.0$), the benefit of correct ordering becomes clearly visible. Recall on the rare minority class rises by $3.1$ percentage points ($p{<}10^{-4}$), and this is not a trade-off against the other classes. Overall accuracy rises too, by a smaller $0.7$ points ($p{<}10^{-4}$). Ordering guarantees are not useless here. They were simply masked by an averaging metric, not absent.

\subsection{Testing the same intervention on interchangeable experts}

We apply the same intervention to the MoE in study (v), deploying each expert alone, with or without additional training. Correctly selecting the initially advantaged expert changes accuracy by only $0.1$--$0.2$ percentage points. Unlike the distinct class targets in study (iv), all six experts share the same architecture and training data, so the initial advantage disappears as training proceeds and their performance converges. The experts therefore become effectively interchangeable, leaving little downstream value for correct ordering to capture.

\subsection{Manipulating interchangeability directly}

To test the interchangeability explanation directly, we confine each expert to a fixed random subset of the 64 input pixels. The advantaged expert $e_{\text{true}}$ retains all pixels, while every other expert loses a fraction denoted \emph{gap}. Masked weights are zero at initialization and their gradients remain masked, making the deficit permanent. $\mathrm{gap}{=}0$ exactly reproduces study (v). For each trained system, we act on every candidate in turn rather than leaving ordering to the gate, balancing the cells by construction and isolating whether choosing the wrong expert has a downstream cost. The position of $e_{\text{true}}$ cycles across seeds.

\begin{table}[h]
\caption{Ordering payoff as a function of candidate interchangeability: invest in the chosen expert, then deploy it. Results are paired within run. 25 seeds per cell, floors pooled ($n{=}50$). $d_z$ is the paired effect size relative to its across-seed variation. Values around 1 or larger are conventionally large.}
\label{tab:fungibility}
\centering
\tablecaptionspace
\footnotesize
\begin{tabular}{@{}lrrrr@{}}
\toprule
gap & mean diff & 95\% CI & $p$ & $d_z$ \\
\midrule
$0.00$ & $-0.0029$ & $[-0.0056,-0.0003]$ & $0.042$ & $-0.30$ \\
$0.05$ & $+0.0069$ & $[+0.0035,+0.0103]$ & $0.0003$ & $+0.55$ \\
$0.10$ & $+0.0144$ & $[+0.0108,+0.0180]$ & $<10^{-4}$ & $+1.08$ \\
$0.20$ & $+0.0358$ & $[+0.0314,+0.0404]$ & $<10^{-4}$ & $+2.20$ \\
$0.35$ & $+0.0756$ & $[+0.0704,+0.0810]$ & $<10^{-4}$ & $+3.88$ \\
$0.50$ & $+0.1414$ & $[+0.1359,+0.1470]$ & $<10^{-4}$ & $+7.04$ \\
\bottomrule
\end{tabular}
\end{table}

Across all $300$ runs the dose-response is slope $+0.280$ (SE $0.006$), $r{=}0.940$, $p{=}1.5\times10^{-141}$ (Spearman $\rho{=}0.909$). Deploy-only scoring gives the same shape (slope $+0.277$, $r{=}0.870$).

Three checks clarify the result. First, under the frozen-gate ensemble the same contrast is flat across gap values (slope $+0.0012$, $p{=}0.49$), showing that the extra training matters only when the selection is actually used. Second, switching the safety floor on or off does not change the ordering effect. At $\mathrm{gap}{=}0.2$, for example, the effect is $+0.0362$ without the floor and $+0.0354$ with it, with similarly close agreement at every other gap. Safety and ordering therefore contribute independently. Third, the informative result is the transition rather than the extreme. The effect is absent at $\mathrm{gap}{=}0$ and becomes detectable at $\mathrm{gap}{=}0.05$, well before the nearly definitional large effect at $\mathrm{gap}{=}0.5$.

\subsection{Scope and reproducibility}

These experiments establish when correct ordering has downstream value. They do not show that residual entropy is the only or best way to obtain that ordering, nor do they retest whether the gate identifies the correct candidate (studies (iv)--(vi)). Instead, they explain the earlier null that correct ordering has little downstream consequence when averaging masks the selection or when the available candidates are effectively interchangeable. Once the selected candidate is acted upon directly and the candidates are made non-interchangeable, the ordering payoff emerges.

The supplementary scripts are \path{selection_payoff_factorial.py} for the direct-consumption control, \path{moe_selection_payoff_factorial.py} for the matched MoE intervention, and \path{moe_fungibility_dose_response.py} for the gap manipulation, run with 50, 50, and 25 seeds, respectively. They reuse functions from the study (iv)--(vi) implementations and are standalone-runnable with public packages and scikit-learn's digits dataset.

In summary, informativeness-aligned ordering produces downstream value when candidates are sufficiently non-interchangeable and the selected candidate is acted upon directly. Ensemble averaging or candidate redundancy masks that benefit, independently of the safety guarantee.

\section{Full Detail: Real-Pipeline Ablation (Study vii) and Downstream Case Study}
\label{app:downstream}

\subsection{Deriving the real pipeline's inner-loop scores}
\label{app:downstream-derivation}

Section~\ref{sec:case-study}'s upstream model instantiates Section~\ref{sec:setup}'s $v_{t,k}(\beta_t)$ concretely, not by analogy. Its Bayesian dynamic-convolution layer maintains a per-position variational weight variance $w_{\sigma,i,p,q}^2$ (channel $i$, kernel position $(p,q)$), driven by two distinct, real, measured context-derived precision statistics per position: $\Lambda_{t,i,p,q}$ (context activation magnitude plus spatial-gradient/edge energy) and $\Lambda_{\mathrm{ctx},t,i,p,q}$ (magnitude alone). For fixed outer bandwidth $\beta_t$, the inner loop sets $w_{\sigma,i,p,q}^2$ to its closed-form stationary point $w_{\sigma,i,p,q}^2=\beta_t/(\beta_t+\Lambda_{t,i,p,q})$, and the epistemic signal consumed downstream is $\mathrm{unc\_A}_t=\sum_k w_{\sigma,t,k}^2\,\Lambda_{\mathrm{ctx},t,k}=\sum_k \Lambda_{\mathrm{ctx},t,k}\,\beta_t/(\beta_t+\Lambda_{t,k})$---exactly Section~\ref{sec:setup}'s $v_{t,k}(\beta_t)=c_{t,k}\beta_t/(\beta_t+a_{t,k})$ with $k=(i,p,q)$, $a_{t,k}=\Lambda_{t,k}$, and $c_{t,k}=\Lambda_{\mathrm{ctx},t,k}$. This is what study~(vii) and the downstream case study below ablate and consume. The assumptions used in Propositions~1--2 hold exactly in this pipeline. Once $\Lambda$ and $\Lambda_{\mathrm{ctx}}$ are fixed, $c_{t,k}$ and $a_{t,k}$ are positive and do not depend on $\beta_t$. Both quantities are computed from the attention-pooled context before $\beta_t$ enters the graph. The production setting \texttt{stop\_grad\_backbone=True} explicitly detaches this context, as verified in the production code. Propositions~1--2 therefore apply directly. Corollaries~3--4 require additional conditions on the converged values, so we evaluate those conditions directly below rather than assuming them.

Unlike studies (v)--(vi) (Algorithm~\ref{alg:moedetail}), $(\gamma,\mu)$ here are updated by true backpropagation through a per-epoch bilevel inner loop, not by Adam with a finite-difference gradient on a semi-frozen surrogate. Algorithm~4 presents this inner and outer training loop, followed by the downstream evaluation, in the exact order used by the production code.

\begin{algbox}{4}{Study (vii)'s exact computational order, real production pipeline (contrast with Algorithms~\ref{alg:phase4detail}--\ref{alg:moedetail}: gradients reach $(\gamma,\mu)$ by true backpropagation through an unrolled inner loop, not finite-difference on a frozen or semi-frozen surrogate). \label{alg:v9detail}}
\textbf{\large Inner loop --- per-window closed-form solution (evaluated every time $L_{\text{outer}}$ is needed)}
\vspace{2pt}\hrule\vspace{4pt}

\algln $t_{\text{norm}}\gets w/(T_{\text{windows}}{-}1)$ for window $w=0,\dots,4$; \ $\beta_w \gets \beta_{\text{base}}\exp(-\gamma(t_{\text{norm}}-\mu)^2)$ \hfill\textit{// Gaussian bandwidth profile, the same $\beta_t(\gamma,\mu)$ role as studies (iv)--(vi)}

\algln forward pass: model$(x_w,\beta_w)\to$ ctx (attention-pooled context); \textbf{if} stop\_grad\_backbone (production default): ctx $\gets$ ctx.detach()

\algln $(\Lambda_w,\Lambda_{\text{ctx},w}) \gets$ \texttt{compute\_lambda}(ctx) \hfill\textit{// $\Lambda$: local MSE + spatial-gradient/edge energy per position $(i,p,q)$; $\Lambda_{\text{ctx}}$: MSE term alone}

\algln $(\Lambda_w,\Lambda_{\text{ctx},w}) \gets$ rescale both by detached factor $\beta_{\text{base}}/\overline{\Lambda_w}$ \hfill\textit{// \texttt{scale\_match\_lambda}: matches $\Lambda$'s scale to $\beta_{\text{base}}$, gradient-blocked}

\algln $w^2_{\sigma,w,i,p,q} \gets \beta_w/(\beta_w+\Lambda_{w,i,p,q})$ \hfill\textit{// \texttt{compute\_w\_sigma}: closed-form stationary point, exact, not approximated}

\algln $\text{unc\_A}_w \gets \sum_{i,p,q} w^2_{\sigma,w,i,p,q}\,\Lambda_{\text{ctx},w,i,p,q} = \sum_k \Lambda_{\text{ctx},w,k}\,\beta_w/(\beta_w+\Lambda_{w,k})$ \hfill\textit{// exactly $v_{t,k}(\beta_t)$ (\S\ref{app:downstream-derivation}), summed over $k{=}(i,p,q)$}

\algln \emph{(consequence of line 2's detach: only the explicit $\beta_w$ factor above carries gradient to $(\gamma,\mu)$ -- $\Lambda,\Lambda_{\text{ctx}}$ never do, the code-level realization of Propositions~1--2's shared hypothesis)}

\algln across all 5 windows: $p_A\gets\text{unc\_A}/\textstyle\sum\text{unc\_A}$, $p_\beta\gets\beta/\textstyle\sum\beta$; \ $L_{\text{outer}} \gets H(p_A)-H(p_\beta)$ \hfill\textit{// \texttt{residual\_entropy\_loss}; the only loss the production module implements}

\vspace{6pt}\hrule\vspace{6pt}
\textbf{\large Outer loop --- per-epoch training loop (300 epochs, one full run per $\mu_{\text{init}}\in\{0.1,0.5,0.9\}$)}
\vspace{2pt}\hrule\vspace{4pt}

\algln \textbf{if} epoch $<50$ (\texttt{OUTER\_WARMUP}): one Adam step on $\theta$ (model weights) per batch, minimizing $L_{\text{data}}=L_{\text{task}}+\hat\beta\cdot\mathrm{KL}(\cdot)$ with $\hat\beta$ from \emph{detached} $(\gamma,\mu)$ \hfill\textit{// backbone shaped; $(\gamma,\mu)$ untouched}

\algln \textbf{elif} \texttt{higher} library available (true bilevel path): collect one representative batch per window; build a differentiable functional copy of the model; run \texttt{INNER\_STEPS} virtual gradient steps of the copy on $L_{\text{task}}$, with $\beta_w$ from the \emph{live} $(\gamma,\mu)$ (trajectory stays differentiable); evaluate the inner loop's $L_{\text{outer}}$ on the virtually-updated copy; \texttt{backward()} through the \emph{entire unrolled inner trajectory} back to $(\gamma,\mu)$ only \hfill\textit{// a true meta-gradient, not finite-difference}

\algln \textbf{else} (fallback, library unavailable): one plain forward pass of the real (non-virtual) model; evaluate the same $L_{\text{outer}}$; backpropagate; gradient-clip; step both optimizers---but the inner loop's detach means no gradient reaches $\theta$ from $L_{\text{outer}}$ here, so only $(\gamma,\mu)$ actually move

\algln clamp $\mu\in[0,1]$, $\gamma\in[0.01,20]$ \hfill\textit{// after either outer branch above}

\algln ordinary $\theta$ update (every epoch, after the outer step attempt): one supervised pass over the full loader, $\beta$ again from \emph{detached} $(\gamma,\mu)$, Adam step on $L_{\text{data}}$; step cosine LR schedule

\algln \textbf{return} checkpoint $(\theta^*,\gamma^*,\mu^*)$; no selection rule across the 3---reproducibility (identical convergence) is the property tested (Fig.~\ref{fig:ablation})

\vspace{6pt}\hrule\vspace{6pt}
\textbf{\large Downstream case study --- inference only (\texttt{allwindow\_eval.py}, \texttt{crossspecies\_mcbaseline\_eval.py}); no new $(\gamma,\mu)$ optimization}
\vspace{2pt}\hrule\vspace{4pt}

\algln evaluate several \emph{already-trained} downstream-predictor checkpoints (uniform baseline; oversampling arms, one built from $\mu^*$'s selected window, one from a standard MC-dropout/BALD \texttt{mc\_var} baseline) at all 5 windows on held-out data

\algln \textbf{return} per-arm, per-window MSE (Tables~\ref{tab:allwindow}--\ref{tab:downstream}) \hfill\textit{// tests whether $\mu^*$'s selection signal helps the downstream predictor, not addressed by Propositions~1--2 themselves}
\end{algbox}

\subsection{Real-pipeline ablation (study vii)}

\begin{center}
\refstepcounter{figure}
\includegraphics[width=0.78\linewidth]{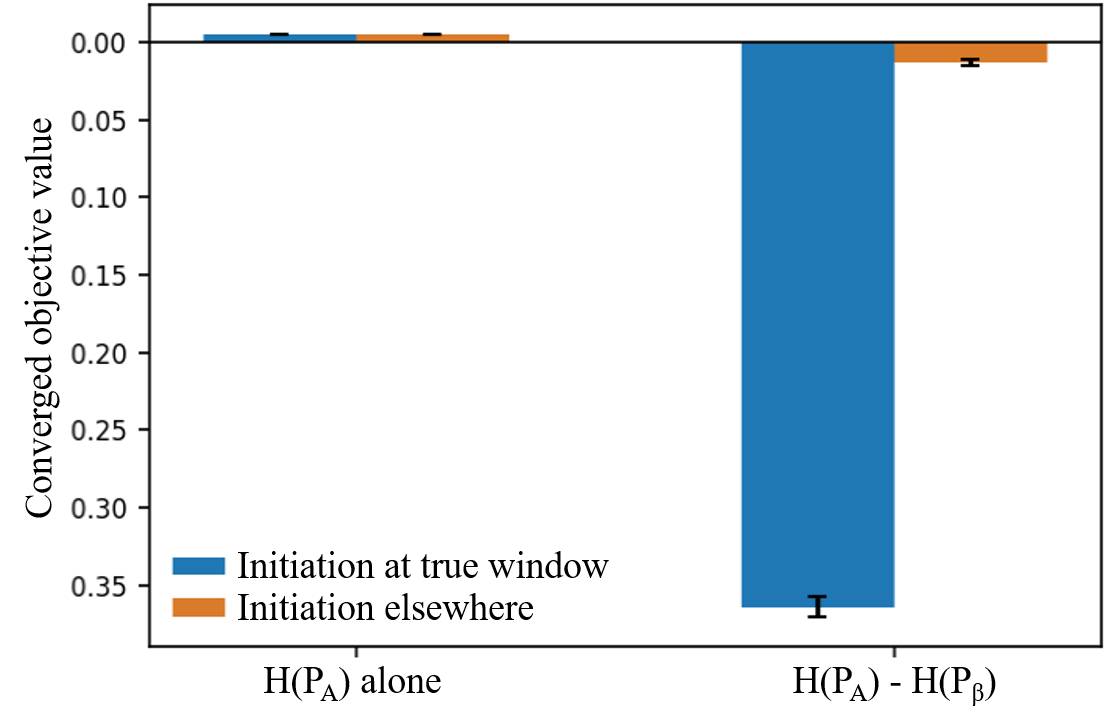}

\small Figure~\thefigure: Before/after ablation on the real upstream pipeline (three $\mu$-initializations, 300 epochs each). Left: $H(p_A)$ alone lets $\gamma$ diverge without bound and the selected window is not reproducible. Right: the residual objective keeps $\gamma$ bounded and all three initializations converge to the same window.
\label{fig:ablation}
\end{center}
\bigskip
This is the reliability result reported in Section~\ref{sec:real-ablation}. It is what makes the downstream case study below trustworthy as a single, reproducible signal rather than an artifact of one lucky initialization. Note that the right panel's $\gamma$ settles at an interior value ($\approx$1.3--1.9 across the three runs, never near $0$ or near $\gamma_{\max}$). This is a reproducible interior optimum, not an instance of Proposition~2's Corollary~2 or Corollary~1.

Do Corollaries~3--4 hold here? No. We evaluated their conditions at each of the three converged checkpoints using \texttt{verify\_prop3\_check\_viii.py} with the production model and its actual checkpoints. Corollary~4's premise failed at all three checkpoints ($0/3$), and its sufficient Cauchy Schwarz condition also failed at all three ($0/3$). Studies~(v)--(vi) produced the same outcome (Appendix~\ref{app:moe-fair}), although through a different optimization regime. Here, $\gamma$ converges to an interior value while $\beta$ remains dispersed, rather than approaching the nearly uniform profile of Corollary~2. Despite this, $D=H(p_A)-H(p_\beta)$ is empirically negative at all three checkpoints ($-0.059$, $-0.057$, $-0.047$ for $\mu_{\mathrm{init}}=0.1,0.5,0.9$)---a safety margin neither checkable corollary guarantees here, but consistent with Proposition~2's general identity, which allows $D\leq0$ even where the checkable corollaries' sufficient condition is not established.

\subsection{Downstream case study}

The headline results are in the main text. Table~\ref{tab:allwindow} shows \texttt{sweetspot} ($\mu^\ast$) beats every other arm pooled across all five windows and both evaluation sources (MSE $92.4$ vs.\ $104.0$ for \texttt{uniform} and $106.5$ for \texttt{mc\_var}), and Table~\ref{tab:downstream} below shows this holds natively on Bacillus ($\sim$19--20\% improvement over \texttt{mc\_var}, $66/90$ pairwise wins across 5 seeds) but weakens and reverses as domain distance grows (Tetraselmis: small edge; HeLa: \texttt{mc\_var} wins).

The benefit is strongest in the native domain. Relative to \texttt{uniform}, it weakens as domain distance grows. Relative to \texttt{mc\_var}, it reverses on HeLa. We present this pattern as a boundary of the method rather than as evidence of uniform success. The purpose of this experiment is to demonstrate real downstream consequences, not to claim state of the art video prediction.

\begin{table}[!htbp]
\caption{Downstream case study: does the correctly-identified sweet-spot window improve a separately-developed video predictor?}
\label{tab:downstream}
\centering
\tablecaptionspace
\footnotesize
\begin{tabularx}{\linewidth}{@{}p{0.26\linewidth}p{0.19\linewidth}X@{}}
\toprule
Domain & Domain distance & Result \\
\midrule
\textit{Bacillus} (native) & native & $\sim$19--20\% MSE improvement over \texttt{mc\_var}; 66/90 pairwise wins (5-seed re-validation); Table~\ref{tab:allwindow} \\
\textit{Tetraselmis} sp.\ (microalga) & diff.\ species, same modality & unanimous vs.\ \texttt{uniform} (3/3 seeds, 6/6 videos); only $+2.5\%$, 4/6 vs.\ \texttt{mc\_var} (1 seed) \\
DIC-C2DH-HeLa (mammalian) & diff.\ organism \& modality & mixed vs.\ \texttt{uniform} (2/3, n.s., $p{=}0.36$); \texttt{mc\_var} wins vs.\ \texttt{sweetspot}, $-5.2\%$, 1/6 (1 seed) \\
\bottomrule
\end{tabularx}
\end{table}
\FloatBarrier

Table~\ref{tab:downstream-tiera} below adds one more piece of evidence not shown in the main text: whether \texttt{sweetspot}'s advantage on Bacillus is broad or narrowly confined to its own oversampled window.

On this subset, \texttt{sweetspot} outperforms \texttt{uniform} at three of the five windows (W2, W4, and W5). At its target window W3, the two methods are nearly tied ($221.3$ vs.\ $219.6$). The improvement is therefore not confined to the oversampled window. Mean MSE also favors \texttt{sweetspot} on this subset ($179.2$ vs.\ $189.9$), matching the direction of the pooled result in Table~\ref{tab:allwindow} ($92.4$ vs.\ $104.0$). The numerical values differ because Table~\ref{tab:allwindow} additionally includes evaluations on two real videos.

\begin{table}[!htbp]
\caption{Study (vii) downstream results on the Bacillus held out synthetic subset (Tier A), reporting per window MSE for three arms.}
\label{tab:downstream-tiera}
\centering
\tablecaptionspace
\small
\begin{tabular}{@{}lrrrrr@{}}
\toprule
Arm & W1 & W2 & W3 & W4 & W5 \\
\midrule
\texttt{uniform} & 208.3 & 66.7 & 219.6 & 265.2 & 189.7 \\
\texttt{sweetspot} ($\mu^\ast{=}$W3) & 231.9 & 53.8 & 221.3 & 217.2 & 171.9 \\
\texttt{mc\_var} ($=$W2) & 246.5 & 80.1 & 200.6 & 309.9 & 150.9 \\
\bottomrule
\end{tabular}
\end{table}
\FloatBarrier

\subsection{Reproducibility}

Table~\ref{tab:allwindow}'s five-arm comparison and Table~\ref{tab:downstream}'s cross-species \texttt{mc\_var} figures come from \texttt{allwindow\_eval.py} (Bacillus, inference only, reusing existing checkpoints) and \texttt{crossspecies\_mcbaseline\_eval.py} (Tetraselmis/HeLa, one new \texttt{mc\_var\_oversample} arm per species). Both are in the supplementary bundle. The factor-sensitivity sweep ($2\times$--$10\times$) and the Bacillus 5-seed re-validation's full per-seed breakdown remain available from the authors upon request, as for the rest of this non-public pipeline.

This appendix is only partially self-contained, unlike Appendices~\ref{app:evidence}--\ref{app:confidence}. Study~(vii)'s ablation harness (\texttt{v9\_outer\_loop\_ablation.py}) is included but not independently re-runnable (it monkeypatches a non-public production module on non-public data). Its real, unedited per-epoch output (\path{ablation_results.json}) is included instead, so Table~\ref{tab:master} row~(vii) and the figure above can be re-analyzed from data alone. The downstream case study's underlying model code, checkpoints, and training data are likewise not included. Its raw JSON outputs (\texttt{allwindow\_eval\_results.json}, \texttt{mc\_baseline\_results.json}$\times 3$ species) are, so Tables~\ref{tab:allwindow}--\ref{tab:downstream} can be re-verified from data alone. Everything else is available from the authors upon request. The DIC-C2DH-HeLa dataset itself is separately public.

\end{document}